\documentclass[10pt,twocolumn,letterpaper]{article}

\usepackage{iccv}
\usepackage{times}
\usepackage{epsfig}
\usepackage{graphicx}
\usepackage{amsmath}
\usepackage{amssymb}

\usepackage{color}
\usepackage{amsmath,amssymb,amsfonts}
\usepackage{multirow}
\usepackage{diagbox}
\usepackage{bbding}
\usepackage{url}
\usepackage{subcaption}
\usepackage{booktabs}

\usepackage[breaklinks=true,bookmarks=false]{hyperref}

\iccvfinalcopy % *** Uncomment this line for the final submission

\def\iccvPaperID{****} % *** Enter the ICCV Paper ID here

\ificcvfinal\fi

\begin{document}

%%%%%%%%% TITLE
\title{Rethinking Rotated Object Detection with Gaussian Wasserstein Distance Loss}

\author{Xue Yang$^{1,2,}$\thanks{Work done during an internship at Huawei Inc.}, Junchi Yan$^{1,2,}$\thanks{Corresponding author is Junchi Yan.}, Qi Ming$^{4}$, Wentao Wang$^{1}$, Xiaopeng Zhang$^{3}$, Qi Tian$^{3}$\\
$^{1}$Department of Computer Science and Engineering, Shanghai Jiao Tong University\\
$^{2}$MoE Key Lab of Artificial Intelligence, AI Institute, Shanghai Jiao Tong University \\
$^{3}$Huawei Inc. \quad
$^{4}$School of Automation, Beijing Institute of Technology \\
{\tt\small yangxue-2019-sjtu@sjtu.edu.cn}
}

\maketitle
\ificcvfinal\thispagestyle{empty}\fi

%%%%%%%%% ABSTRACT
\begin{abstract}
   Boundary discontinuity and its inconsistency to the final detection metric have been the bottleneck for rotating detection regression loss design. In this paper, we propose a novel regression loss based on Gaussian Wasserstein distance as a fundamental approach to solve the problem. Specifically, the rotated bounding box is converted to a 2-D Gaussian distribution, which enables to approximate the indifferentiable rotational IoU induced loss by the Gaussian Wasserstein distance (GWD) which can be learned efficiently by gradient back-propagation. GWD can still be informative for learning even there is no overlapping between two rotating bounding boxes which is often the case for small object detection. Thanks to its three unique properties, GWD can also elegantly solve the boundary discontinuity and square-like problem regardless how the bounding box is defined. Experiments on five datasets using different detectors show the effectiveness of our approach. Codes are made public available\footnote{https://github.com/yangxue0827/RotationDetection}\footnote{https://github.com/open-mmlab/mmrotate}.
\end{abstract}

%%%%%%%%% BODY TEXT
\section{Introduction}

Arbitrary-oriented objects are ubiquitous for detection across visual datasets, such as aerial images~\cite{yang2018automatic, azimi2018towards, ding2018learning, yang2019scrdet}, scene text~\cite{zhou2017east, liu2018fots, jiang2017r2cnn, ma2018arbitrary, liao2018rotation}, faces~\cite{shi2018real} and 3D objects~\cite{zheng2020rotation}, retail scenes~\cite{chen2020piou, pan2020dynamic}, etc. Compared with the large literature on horizontal object detection~\cite{girshick2015fast,ren2015faster,lin2017feature,lin2017focal,dai2016r}, research in oriented object detection is relatively in its earlier stage, with many open problems to solve.

\begin{figure}[!tb]
	\centering
	\begin{subfigure}{.23\textwidth}
		\centering	
		\includegraphics[width=.98\linewidth]{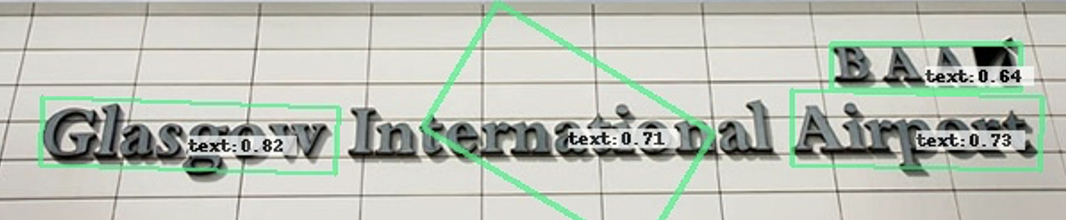}
		\label{fig:compare_vis_1}
	\end{subfigure}
	\begin{subfigure}{.23\textwidth}
		\centering	
		\includegraphics[width=.98\linewidth]{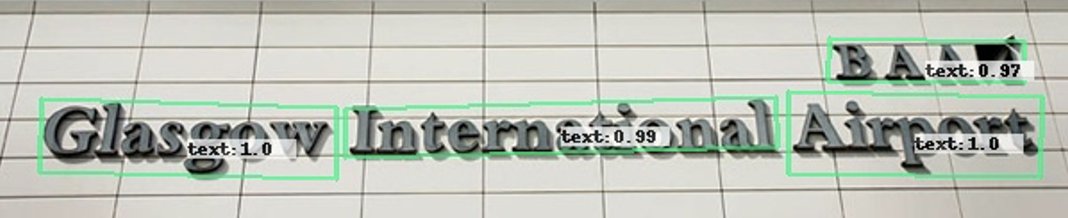}
		\label{fig:compare_vis_2}
	\end{subfigure}\\
	\begin{subfigure}{.23\textwidth}
		\centering	
		\includegraphics[width=.98\linewidth]{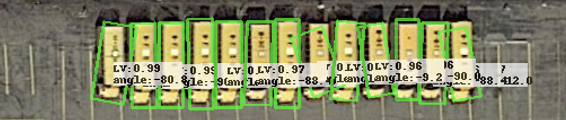}
		\label{fig:compare_vis_3}
	\end{subfigure}
	\begin{subfigure}{.23\textwidth}
		\centering	
		\includegraphics[width=.98\linewidth]{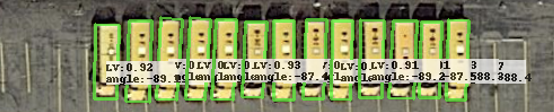}
		\label{fig:compare_vis_4}
	\end{subfigure}\\
	\begin{subfigure}{.23\textwidth}
		\centering	
		\includegraphics[width=.98\linewidth]{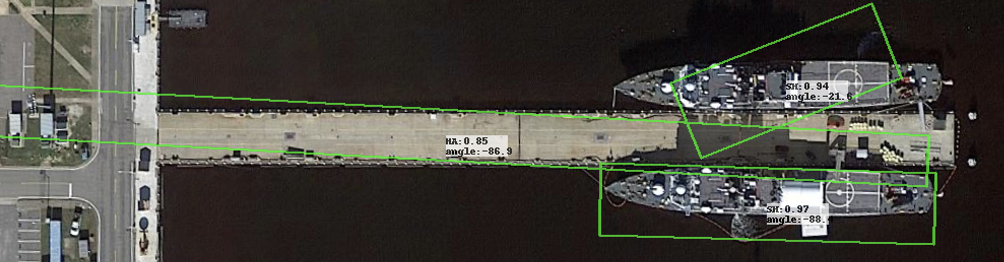}
		\label{fig:compare_vis_5}
	\end{subfigure}
	\begin{subfigure}{.23\textwidth}
		\centering	
		\includegraphics[width=.98\linewidth]{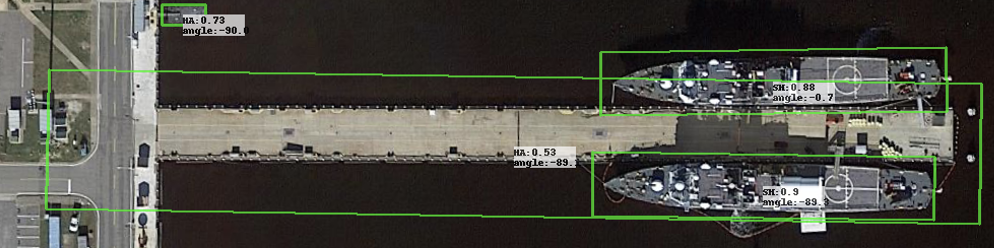}
		\label{fig:compare_vis_6}
	\end{subfigure}
	\caption{Comparison of the detection results between Smooth L1 loss-based (left) and the proposed GWD-based (right) detector.}
	\label{fig:compare_vis}
\end{figure}

The dominant line of works~\cite{azimi2018towards, ding2018learning, yang2019scrdet, yang2021r3det} take a regression methodology to predict the rotation angle, which has achieved state-of-the-art performance. However, compared with traditional horizontal detectors, the angle regression model will bring new issues, as summarized as follows: i) the inconsistency between metric and loss, ii) boundary discontinuity, and iii) square-like problem. In fact, these issues remain open without a unified solution, and they can largely hurt the final performance especially at the boundary position, as shown in the left of Fig.~\ref{fig:compare_vis}. In this paper, we use a two-dimensional Gaussian distribution to model an arbitrary-oriented bounding box for object detection, and approximate the indifferentiable rotational Intersection over Union (IoU) induced loss between two boxes by calculating their Gaussian Wasserstein Distance (GWD)~\cite{Wasserstein2010djalil}.

GWD elegantly aligns model learning with the final detection accuracy metric, which has been a bottleneck and not achieved in existing rotation detectors. Our GWD based detectors are  immune from both boundary discontinuity and square-like problem, and this immunity is independent with how the bounding box protocol is defined, as shown on the right  of Fig.~\ref{fig:compare_vis}. The highlights of this paper are four-folds:

i) We summarize three flaws in state-of-the-art  rotation detectors, i.e. inconsistency between metric and loss, boundary discontinuity, and square-like problem, due to their regression based angle prediction nature.

ii) We propose to model the rotating bounding box distance by Gaussian Wasserstein Distance (GWD) which leads to an approximate and differentiable IoU induced loss. It resolves the loss inconsistency by aligning model learning with accuracy metric and thus naturally improves the model. %To our best knowledge, this is the first work that aligns loss with performance metric for rotation detection.

iii) Our GWD-based loss can elegantly resolve boundary discontinuity and square-like problem, regardless how the rotating bounding box is defined. In contrast, the design of most peer works \cite{yang2020arbitrary,yang2020dense} are coupled with the  parameterization of bounding box.

iv) Extensive experimental results on five public datasets and two popular detectors show the effectiveness of our approach. The source codes \cite{yang2021alpharotate,zhou2022mmrotate} are made public available.

%-------------------------------------------------------------------------

\section{Related Work}
\label{sec:related}
In this paper, we mainly discuss the related work on rotating object detection. Readers are referred to \cite{girshick2015fast,ren2015faster,lin2017feature,lin2017focal} for more comprehensive literature review on horizontal object detection.

\textbf{Rotated object detection.} As an emerging direction, advance in this area try to extend classical horizontal detectors to the rotation case by adopting the rotated bounding boxes. Compared with the few works~\cite{yang2020arbitrary} that treat the rotation detection tasks an angle classification problem, regression based detectors still dominate which have been applied in different applications. For aerial images, ICN \cite{azimi2018towards}, ROI-Transformer \cite{ding2018learning}, SCRDet \cite{yang2019scrdet} and Gliding Vertex \cite{xu2020gliding} are two-stage representative methods whose pipeline comprises of object localization and classification, while DRN \cite{pan2020dynamic}, R$^3$Det \cite{yang2021r3det} and RSDet \cite{qian2021learning} are single-stage methods. For scene text detection, RRPN \cite{ma2018arbitrary} employ rotated RPN to generate rotated proposals and further perform rotated bounding box regression. TextBoxes++ \cite{liao2018textboxes++} adopts vertex regression on SSD. RRD \cite{liao2018rotation} further improves TextBoxes++ by decoupling classification and bounding box regression on rotation-invariant and rotation sensitive features, respectively. We discuss the specific challenges in existing regressors for rotation detection.

\textbf{Boundary discontinuity and square-like problems.} 
Due to the periodicity of angle parameters and the diversity of bounding box definitions, regression-based rotation detectors often suffer from boundary discontinuity and square-like problem. Many existing methods try to solve part of the above problems from different perspectives. For instance, SCRDet~\cite{yang2019scrdet} and RSDet~\cite{qian2021learning} propose IoU-smooth L1 loss and modulated loss to smooth the the boundary loss jump. CSL~\cite{yang2020arbitrary} transforms angular prediction from a regression problem to a classification one. DCL~\cite{yang2020dense} further solves square-like object detection problem introduced by the long edge definition, which refers to rotation insensitivity issue for instances that are approximately in square shape, which will be detailed in Sec.~\ref{sec:revisit}.

\textbf{Approximate differentiable rotating IoU loss.}
It has been shown in classic horizontal detectors that the use of IoU induced loss e.g. GIoU \cite{rezatofighi2019generalized}, DIoU \cite{zheng2020distance} can ensure the consistency of the final detection metric and loss. However, these IoU loss cannot be applied directly in rotation detection because the rotating IoU is indifferentiable. Many efforts have been made to finding an approximate IoU loss for gradient computing. PIoU~\cite{chen2020piou} is realized by simply counting the number of pixels. To tackle the uncertainty of convex caused by rotation, \cite{zheng2020rotation} proposes a projection operation to estimate the intersection area. SCRDet~\cite{yang2019scrdet} combines IoU and smooth L1 loss to develop an IoU-smooth L1 loss, which partly circumvents the need for differentiable rotating IoU loss. 

So far, there exists no truly unified solution to all the above problems which are in fact interleaved to each other. Our method  addresses all these issues in a unified manner. It is also decoupled from the specific definition of bounding box. All these merits make our approach elegant and effective. 

\begin{figure}[!tb]
	\begin{center}
		\includegraphics[width=0.95\linewidth]{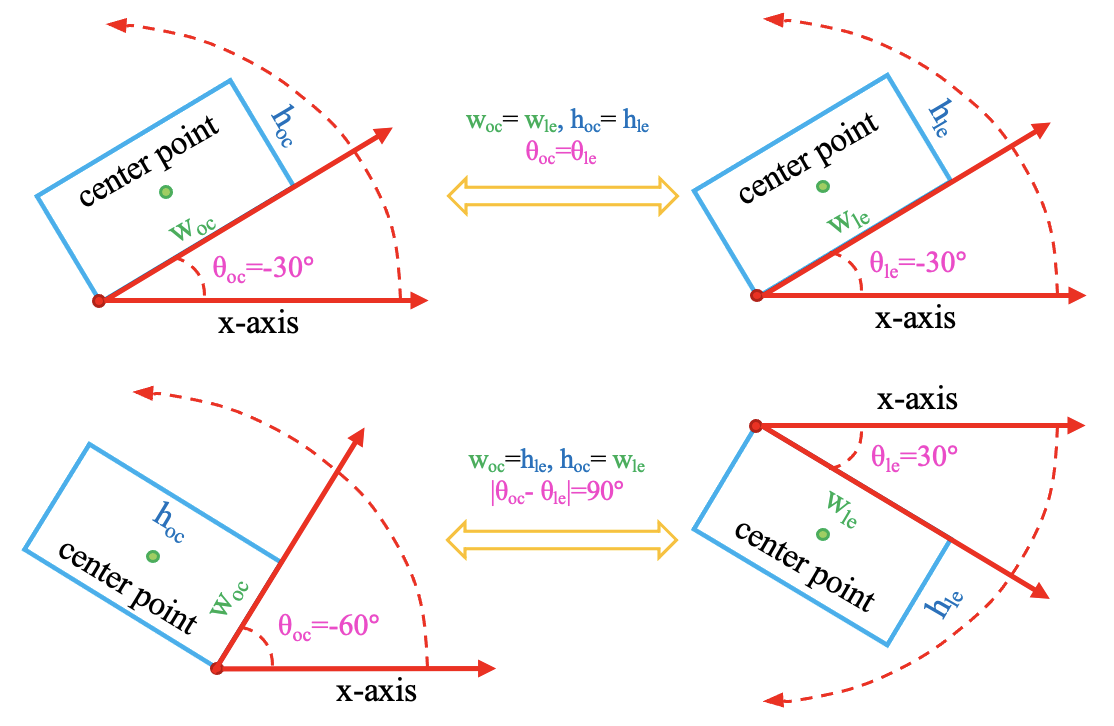}
	\end{center}
	\caption{Two definitions of bounding boxes. \textbf{Left:} OpenCV Definition $D_{oc}$, \textbf{Right:} Long Edge Definition $D_{le}$.}
	\label{fig:definition}
\end{figure}

%-------------------------------------------------------------------------
\section{Rotated Object Regression Detector Revisit}
\label{sec:revisit}
To motivate this work, in this section, we introduce and analyze some deficiencies in state-of-the-art rotating detectors, which are mostly based on angle regression.
%\subsection{The Coupling between Rotating Bounding Box Definition and Detector Design}
\subsection{Bounding Box Definition}

%Recent advances in arbitrary-oriented object detection are mainly driven by adaption of classical object detectors. Among them, angle regression based algorithms~\cite{ding2018learning, yang2019scrdet, yang2022scrdet++, yang2021r3det} dominate, and they still use classic regression losses, e.g. $l_{n}$-norms, defined on parametric representation of two arbitrary-oriented bounding boxes in 2D/3D and improving their IoU values. Although angle regression based algorithms with $l_{n}$-norms loss can achieve satisfactory performance, it is not the optimal combination, and some non-ignorable errors will occur in some scenarios. We will analyze several deficiencies in the above methods in detail.

Fig.~\ref{fig:definition} gives two popular definitions for parameterizing rotating bounding box based angles: OpenCV protocol denoted by $D_{oc}$, and long edge definition by $D_{le}$. Note $\theta \in [-90^\circ, 0^\circ)$ of the former denotes the acute or right angle between the $w_{oc}$ of bounding box and $x$-axis. In contrast,  $\theta \in [-90^\circ, 90^\circ)$ of the latter definition is the angle between the long edge $w_{le}$ of bounding box and $x$-axis. The two kinds of parameterization can be converted to each other:
\begin{equation*}
    \small
        D_{le}(w_{le},h_{le},\theta_{le}) = \\
    	\left\{ \begin{array}{rcl}
    	D_{oc}(w_{oc},h_{oc},\theta_{oc}), & w_{oc} \geq h_{oc} \\ D_{oc}(h_{oc},w_{oc},\theta_{oc} + 90^\circ), & otherwise
    	\end{array}\right.
    	\label{eq:oc2le}
    \end{equation*}
    \begin{equation*}
    \small
        D_{oc}(w_{oc},h_{oc},\theta_{oc}) = \\
    	\left\{ \begin{array}{rcl}
    	D_{le}(w_{le},h_{le},\theta_{le}), & \theta_{le} \in [-90^\circ,0^\circ) \\ D_{le}(h_{le},w_{le},\theta_{le} - 90^\circ), & otherwise
    	\end{array}\right.
    	\label{eq:le2oc}
    \end{equation*}

The main difference refers to the edge and angle $(w, h, \theta)$: when the same bounding box takes different representations by the two definitions, the order of the edges is exchanged and the angle difference is $90^\circ$.

\begin{figure}[!tb]
	\centering
	\begin{subfigure}{.46\textwidth}
		\centering	
		\includegraphics[width=.98\linewidth]{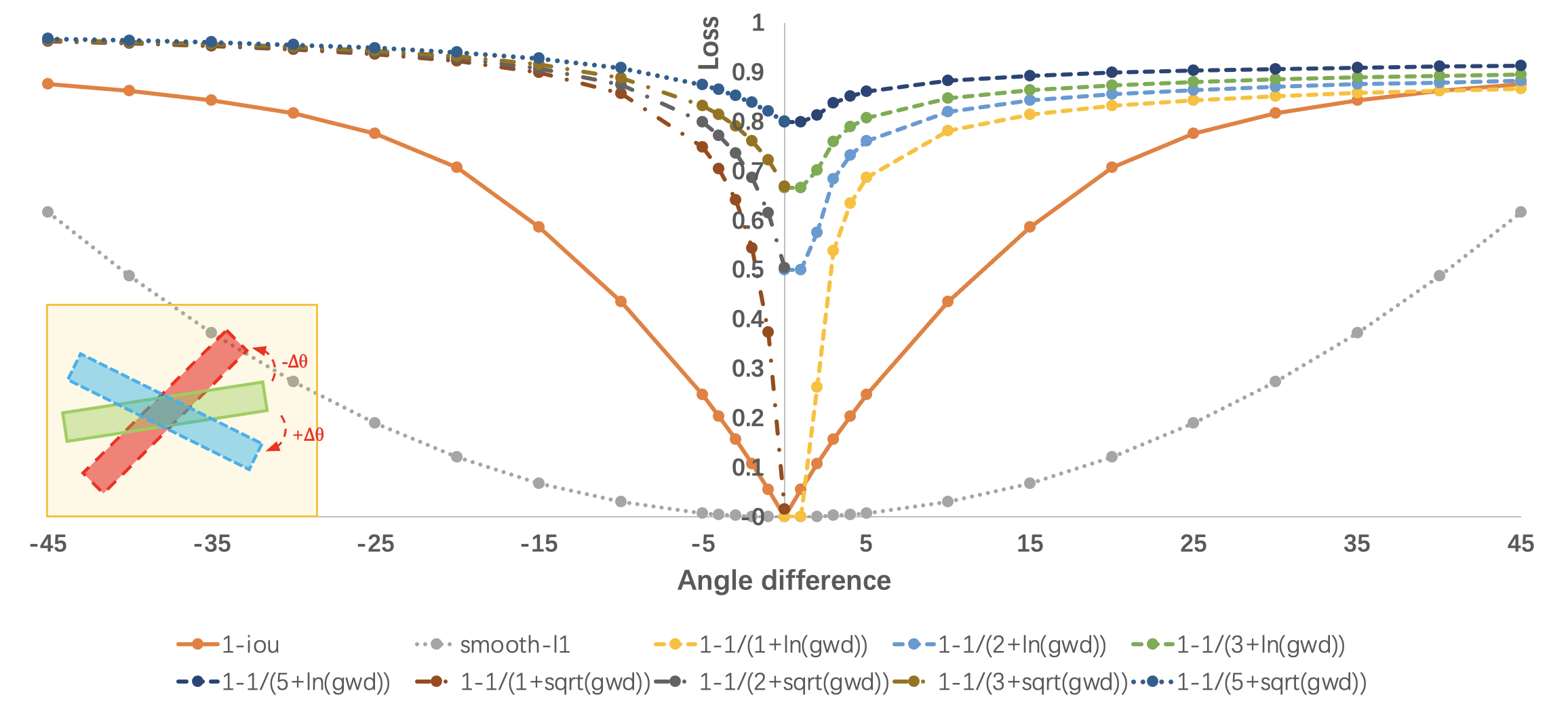}
		\caption{Angle difference case}
		\label{fig:iou_smooth-l1-1}
	\end{subfigure}\\
	\begin{subfigure}{.23\textwidth}
		\centering	
		\includegraphics[width=.98\linewidth]{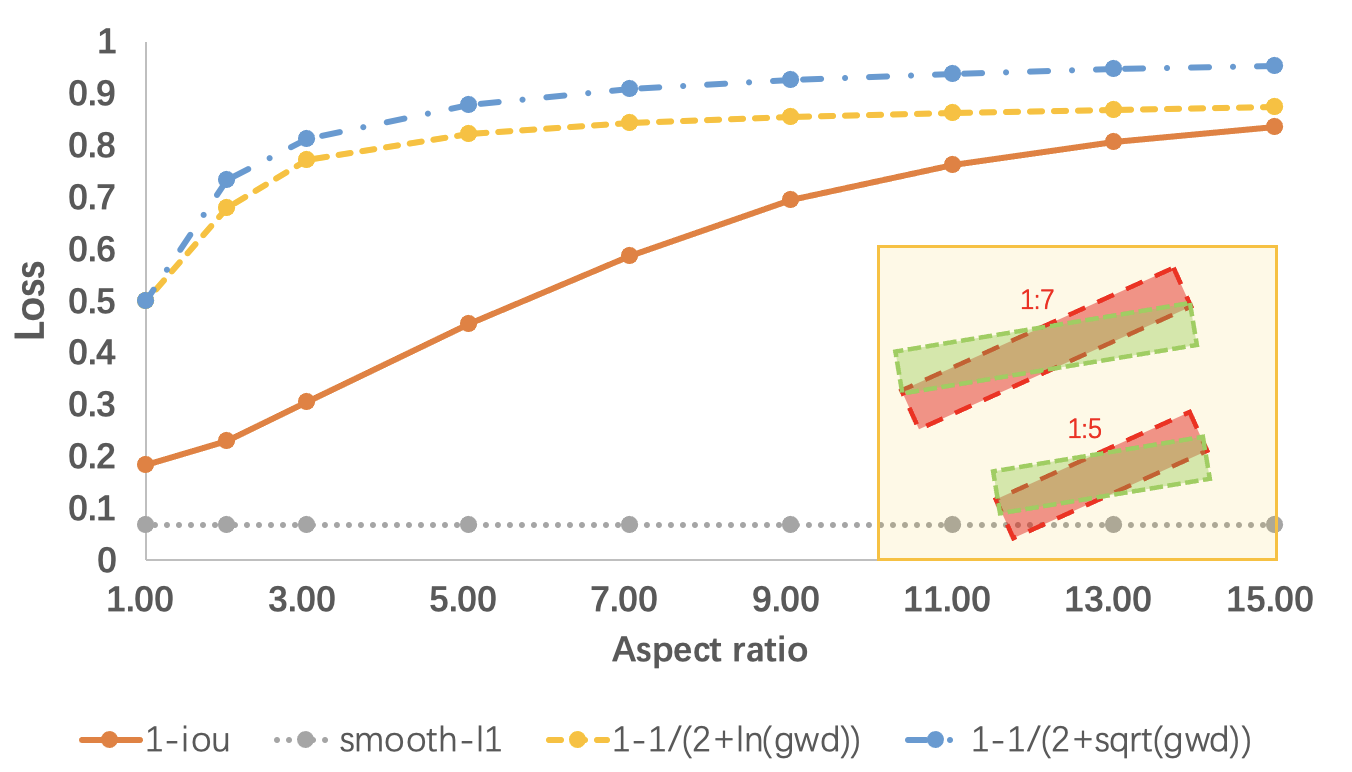}
		\caption{Aspect ratio case}
		\label{fig:iou_smooth-l1-2}
	\end{subfigure}
	\begin{subfigure}{.23\textwidth}
		\centering	
		\includegraphics[width=.98\linewidth]{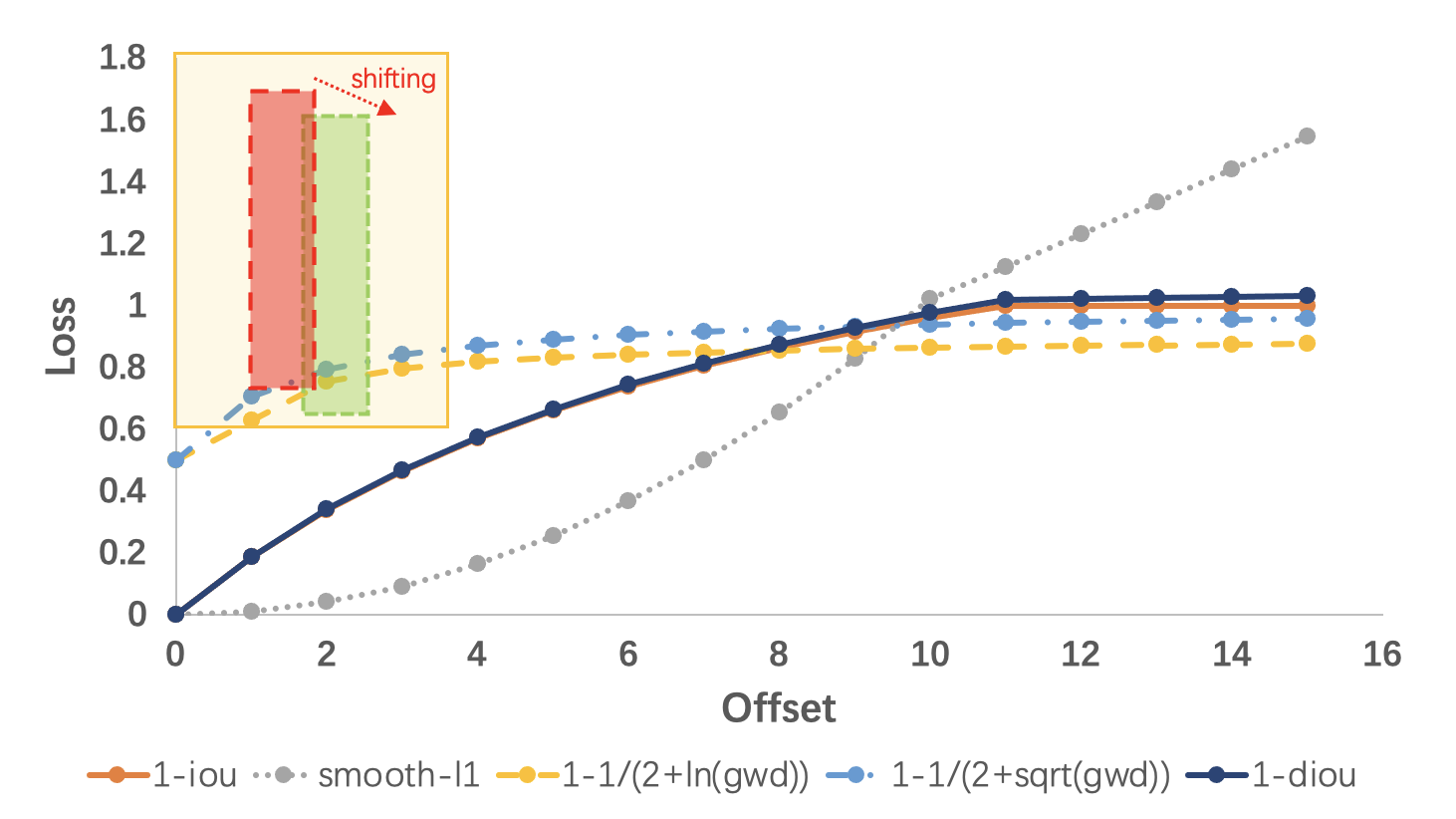}
		\caption{Center shifting case}
		\label{fig:iou_smooth-l1-3}
	\end{subfigure}
	\caption{Behavior comparison of different loss in different cases.}
	\label{fig:iou_smooth-l1}
\end{figure}

In many works, the pipeline design are tightly coupled with the choice of the bounding box definition to avoid specific problems: SCRDet~\cite{yang2019scrdet}, R$^3$Det~\cite{yang2021r3det} are based on $D_{oc}$ to avoid the square-like problem, while CSL~\cite{yang2020arbitrary}, DCL~\cite{yang2020dense} resort to $D_{le}$ to avoid the exchangeability of edges (EoE). %In fact, the effects of different definition are entangled with other factors including the network, datasets, and hyperparameters. There is no elegant solution to decouple the detector design from the definition choice.

\subsection{Inconsistency between Metric and Loss}
Intersection over Union (IoU) has been the standard metric for both horizontal detection and rotation detection. However, there is an inconsistency between the metric and regression loss (e.g. $l_{n}$-norms), that is, a smaller training loss cannot guarantee a higher performance, which has been extensively discussed in horizontal detection \cite{rezatofighi2019generalized, zheng2020distance}. This misalignment becomes more prominent in rotating object detection due to the introduction of angle parameter in regression based models. To illustrate this, we use Fig.~\ref{fig:iou_smooth-l1} to compare IoU induced loss and smooth L1 loss \cite{girshick2015fast}:

\textbf{Case 1:} Fig.~\ref{fig:iou_smooth-l1-1} depicts the relation between angle difference and loss functions. Though they all bear monotonicity, only smooth L1 curve is convex while the others are not.

\textbf{Case 2:} Fig.~\ref{fig:iou_smooth-l1-2} shows the changes of the two loss functions under different aspect ratio conditions. It can be seen that the smooth L1 loss of the two bounding box are constant (mainly from the angle difference), but the IoU loss will change drastically as the aspect ratio varies.

\textbf{Case 3:} Fig.~\ref{fig:iou_smooth-l1-3} explores the impact of center point shifting on different loss functions. Similarly, despite the same monotonicity, there is no high degree of consistency.

\begin{figure*}[!tb]
	\begin{center}
		\includegraphics[width=0.95\linewidth]{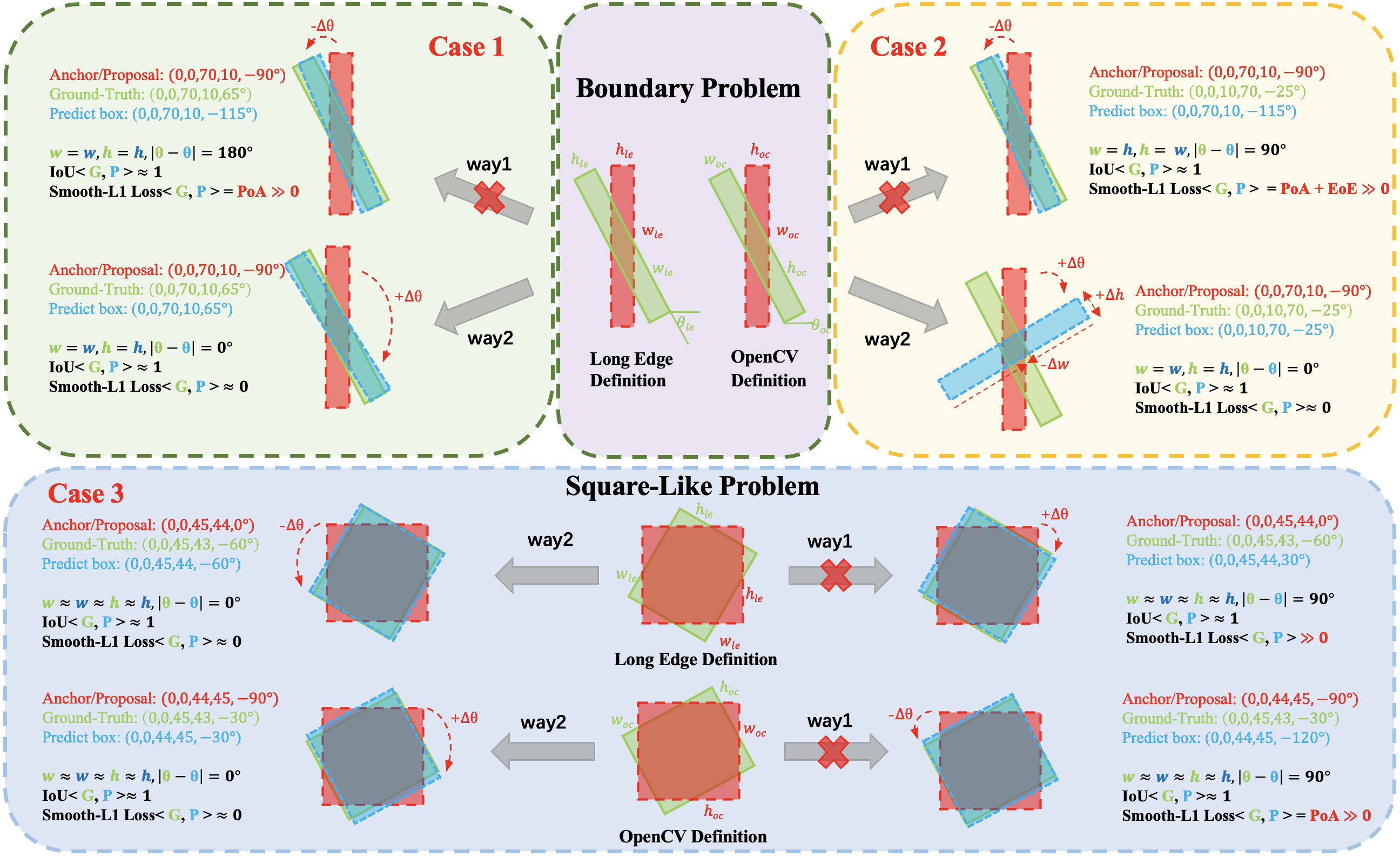}
	\end{center}
	\caption{Boundary discontinuity under two bounding box definitions (top), and illustration of the square-like problem (bottom).}
	\label{fig:problems}
\end{figure*}

Seeing the above flaws of classic smooth L1 loss, IoU-induced loss has become recently popular for horizontal detection e.g. GIoU \cite{rezatofighi2019generalized}, DIoU \cite{zheng2020distance}. It can help fill the gap between metric and regression loss for rotating object detection. However, different from horizontal detection, the IoU of two rotating boxes is indifferentiable for learning. In this paper, we propose a differentiable loss based on Wasserstein distance of two rotating boxes to replace the hard IoU loss. It is worth mentioning that the Wasserstein distance function has some unique properties to solve boundary discontinuity and square-like problem, which will be detailed later.

\subsection{Boundary Discontinuity and Square-Like Problem}
As a standing issue for regression-based rotation detectors, the boundary discontinuity ~\cite{yang2019scrdet,yang2020arbitrary} in general refers to the sharp loss increase at the boundary induced by the angle and edge parameterization. 

Specifically, \textbf{Case 1-2} in Fig.~\ref{fig:problems} summarize the boundary discontinuity. Take \textbf{Case 2} as an example, we assume that there is a red anchor/proposal $\color{red}{(0,0,70,10,-90^\circ)}$ and a green ground truth $\color{green}{(0,0,10,70,-25^\circ)}$ at the boundary position\footnote{The angle of the bounding box is close to the maximum and minimum values of the angle range. For more clearly visualization, the ground truth has been rendered with a larger angle in Fig.~\ref{fig:problems}.}, both of which are defined in OpenCV definition $D_{oc}$. 
The upper right corner of Fig.~\ref{fig:problems} shows two ways to regress from anchor/proposal to ground truth. The \textbf{way1} achieves the goal by only rotating anchor/proposal by an angle counterclockwise, but a very large smooth L1 loss occurs at this time due to the periodicity of angle (PoA) and the exchangeability of edges (EoE). As discussed in CSL \cite{yang2020arbitrary}, this is because the result of the prediction box $\color{blue}{(0,0,70,10,-115^\circ)}$ is outside the defined range. As a result, the model has to make predictions in other complex regression forms, such as rotating anchor/proposal by an large angle clockwise to the blue box while scaling $w$ and $h$ (\textbf{way2} in \textbf{Case 2}). A similar problem (only PoA) also occurs in the long edge definition $D_{le}$, as shown in \textbf{Case 1}. 

In fact, when the predefined anchor/proposal and ground truth are not in the boundary position, \textbf{way1} will not produce a large loss.
Therefore, there exists inconsistency between the boundary position and the non-boundary position regression, which makes the model very confused about in which way it should perform regression. Since non-boundary cases account for the majority, the regression results of models, especially those with weaker learning capacity, are fragile in boundary cases, as shown in the left of Fig.~\ref{fig:compare_vis}.

In addition, there is also a square-like object detection problem in the $D_{le}$-based method~\cite{yang2020dense}. First of all, the $D_{le}$ cannot uniquely define a square bounding box. For square-like objects\footnote{Many instances are in square shape. For instance, two categories of storage-tank (ST) and roundabout (RA) in DOTA dataset.}, $D_{le}$-based method will encounter high IoU but high loss value similar to the boundary discontinuity, as shown by the upper part of \textbf{Case 3} in Fig.~\ref{fig:problems}. In \textbf{way1}, the red anchor/proposal $\color{red}{(0,0,45,44,0^\circ)}$ rotates a small angle clockwise to get the blue prediction box. The IoU of ground truth $\color{green}{(0,0,45,43,-60^\circ)}$ and the prediction box $\color{blue}{(0,0,45,44,30^\circ)}$ is close to 1, but the regression loss is high due to the inconsistency of angle parameters. Therefore, the model will rotate a larger angle counterclockwise to make predictions, as described by \textbf{way2}. The reason for the square-like problem in $D_{le}$-based method is not the above-mentioned PoA and EoE, but the inconsistency of evaluation metric and loss. In contrast, the negative impact of EoE will be weakened when we use $D_{oc}$-based method to detect square-like objects, as shown in the comparison between \textbf{Case 2} and the lower part of \textbf{Case 3}. Therefore, there is no square-like problem in the $D_{oc}$-based method.

\begin{figure}[!tb]
	\begin{center}
		\includegraphics[width=0.95\linewidth]{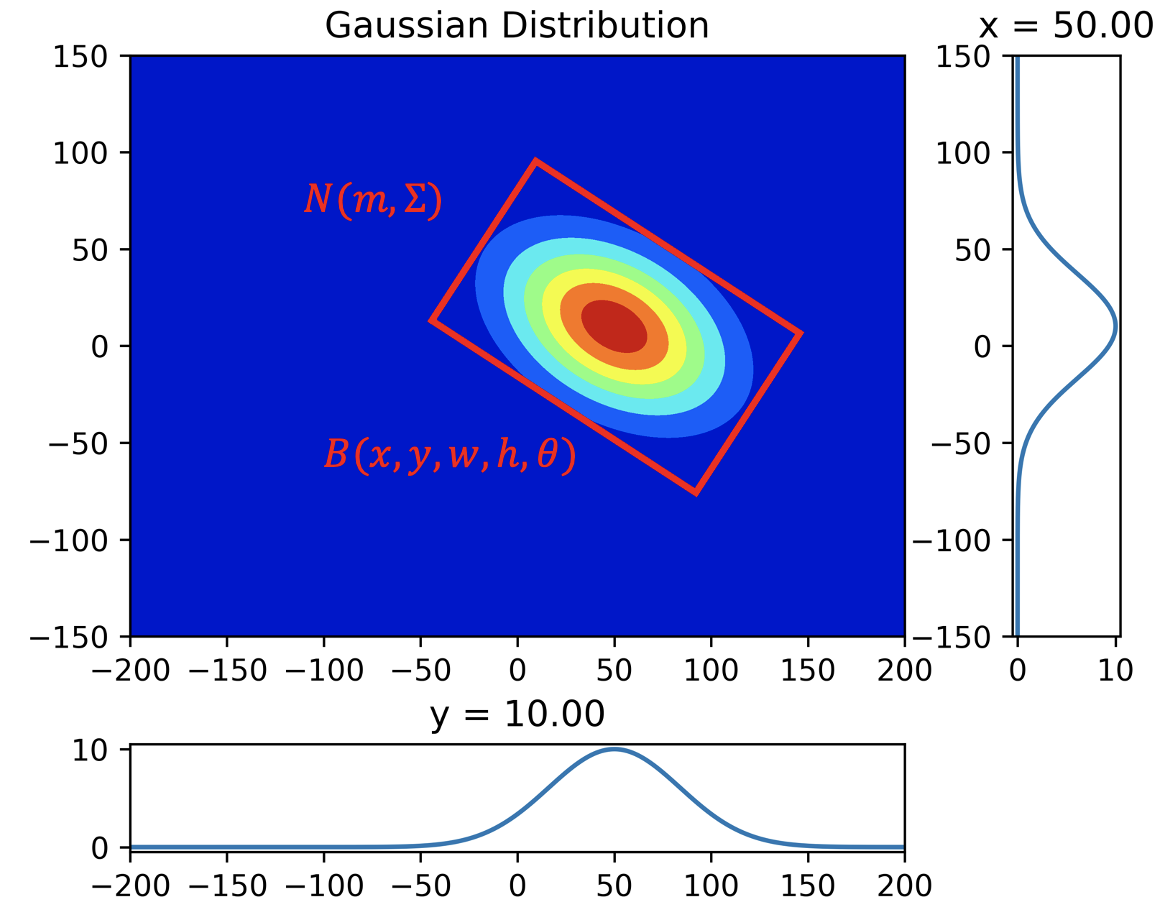}
	\end{center}
	\caption{A schematic diagram of modeling a rotating bounding box by a two-dimensional Gaussian distribution.}
	\label{fig:gpd}
\end{figure}

Recent methods start to address these issues. SCRDet~\cite{yang2019scrdet} combines IoU and smooth L1 loss to propose a IoU-smooth L1 loss, which does not require the rotating IoU being differentiable. It also solves the problem of inconsistency between loss and metric by eliminating the discontinuity of loss at the boundary. However, SCRDet still needs to determine whether the predicted bounding box result conforms to the current bounding box definition method before calculating the IoU. In addition, the gradient direction of IoU-Smooth L1 Loss is still dominated by smooth L1 loss. RSDet~\cite{qian2021learning} devises modulated loss to smooth the loss mutation at the boundary, but it needs to calculate the loss of as many parameter combinations as possible. CSL~\cite{yang2020arbitrary} transforms angular prediction from a regression problem to a classification problem. CSL needs to carefully design their method according to the  bounding box definition ($D_{le}$), and is limited by the  classification granularity with theoretical limitation for high-precision angle prediction. On the basis of CSL, DCL~\cite{yang2020dense} further solves the problem of square-like object detection introduced by $D_{le}$.
%-------------------------------------------------------------------------

\section{The Proposed Method}
\label{sec:method}
%Obviously, the above three issues are not independent, including inconsistency between metric and loss, boundary discontinuity and square-like problem. There is no unified solution has been proposed to solve all the above problems yet. In order to achieve this goal, the new regression loss need to meet the following three requirements: 

In this section we introduce a new rotating object detector whose regression loss fulfills the following requirements:

\textbf{Requirement 1:}  highly consistent with the IoU induced metrics (which  also solves the square-like object problem);

\textbf{Requirement 2:} differentiable allowing for direct learning;

\textbf{Requirement 3:} smooth at angle boundary case.

%It provides a unified solution to all the above problems. 

\subsection{Wasserstein Distance for Rotating Box}
Most of the IoU-based loss can be considered as a distance function. Inspired by this, we propose a new regression loss based on Wasserstein distance. First, we convert a rotating bounding box $\mathcal{B}(x,y,w,h,\theta)$ into a 2-D Gaussian distribution $\mathcal{N}(\mathbf{m},\mathbf{\Sigma})$ (see Fig.~\ref{fig:gpd}) by the following formula:
\begin{equation}
\small
    \begin{aligned}
        \mathbf{\Sigma}^{1/2}=&\mathbf{RSR}^{\top}\\
        =&
        \left(                 
          \begin{array}{cc}   
            \cos{\theta} & -\sin{\theta}\\  
            \sin{\theta} & \cos{\theta}\\  
          \end{array}
        \right)
        \left(                 
          \begin{array}{cc}   
            \frac{w}{2} & 0\\  
            0 & \frac{h}{2}\\  
          \end{array}
        \right)
        \left(                 
          \begin{array}{cc}   
            \cos{\theta} & \sin{\theta}\\  
            -\sin{\theta} & \cos{\theta}\\  
          \end{array}
        \right)\\
        =&
        \left(                 
          \begin{array}{cc}   
            \frac{w}{2}\cos^2{\theta}+\frac{h}{2}\sin^2{\theta} & \frac{w-h}{2}\cos{\theta}\sin{\theta}\\  
            \frac{w-h}{2}\cos{\theta}\sin{\theta} & \frac{w}{2}\sin^2{\theta}+\frac{h}{2}\cos^2{\theta}\\  
          \end{array}
        \right)\\
        \mathbf{m}=&(x,y)
    \end{aligned}
	\label{eq:wd1}
\end{equation}
where $\mathbf{R}$ represents the rotation matrix, and $S$ represents the diagonal matrix of eigenvalues.

The Wasserstein distance $\mathbf{W}$ between two probability measures $\mu$ and $\nu$ on $\mathbb{R}^n$ expressed as~\cite{Wasserstein2010djalil}:
\begin{equation}
\small
    \begin{aligned}
        \mathbf{W}(\mu;\nu):= \inf \mathbb{E}(\lVert \mathbf{X}-\mathbf{Y}\rVert_{2}^{2})^{1/2}
    \end{aligned}
	\label{eq:wd2}
\end{equation}
where the inferior runs over all random vectors $(\mathbf{X},\mathbf{Y})$ of $\mathbb{R}^n\times \mathbb{R}^n$ with $\mathbf{X} \sim \mu$ and $\mathbf{Y} \sim \nu$. It turns out that we have $d:=\mathbf{W}(\mathcal{N}(\mathbf{m}_{1},\mathbf{\Sigma}_{1});\mathcal{N}(\mathbf{m}_{2},\mathbf{\Sigma}_{2}))$ and it writes as:
\begin{equation}
\small
    \begin{aligned}
        d^{2}=\lVert \mathbf{m}_{1}-\mathbf{m}_{2}\rVert_{2}^{2}+\mathbf{Tr}\left(\mathbf{\Sigma}_{1}+\mathbf{\Sigma}_{2}-2(\mathbf{\Sigma}_{1}^{1/2}\mathbf{\Sigma}_{2}\mathbf{\Sigma}_{1}^{1/2})^{1/2}\right)
    \end{aligned}
	\label{eq:wd3}
\end{equation}
This formula has interested several works~\cite{givens1984class,olkin1982distance,knott1984optimal,dowson1982frechet}. Note in particular we have:
\begin{equation}
    \small
    \begin{aligned}
        \mathbf{Tr}\left((\mathbf{\Sigma}_{1}^{1/2}\mathbf{\Sigma}_{2}\mathbf{\Sigma}_{1}^{1/2})^{1/2}\right)=\mathbf{Tr}\left((\mathbf{\Sigma}_{2}^{1/2}\mathbf{\Sigma}_{1}\mathbf{\Sigma}_{2}^{1/2})^{1/2}\right)
    \end{aligned}
	\label{eq:wd4}
\end{equation}
In the commutative case (horizontal detection task)   $\mathbf{\Sigma}_{1}\mathbf{\Sigma}_{2}=\mathbf{\Sigma}_{2}\mathbf{\Sigma}_{1}$, Eq.~\ref{eq:wd3} becomes:
\begin{equation}
\small
    \begin{aligned}
        d^{2}=&\lVert \mathbf{m}_{1}-\mathbf{m}_{2}\rVert_{2}^{2}+\lVert \mathbf{\Sigma}_{1}^{1/2}-\mathbf{\Sigma}_{2}^{1/2} \rVert_{F}^{2}\\
        =&(x_{1}-x_{2})^2+(y_{1}-y_{2})^2+\frac{(w_{1}-w_{2})^2+(h_{1}-h_{2})^2}{4}\\
        =&l_{2}\text{-norm}\left(\left[x_1,y_1,\frac{w_1}{2},\frac{h_1}{2}\right]^\top,\left[x_2,y_2,\frac{w_2}{2},\frac{h_2}{2}\right]^\top\right)
    \end{aligned}
	\label{eq:wd5}
\end{equation}
where $\|\|_F$ is the Frobenius norm. Note that both boxes are horizontal here, and Eq.~\ref{eq:wd5} is approximately equivalent to the $l_{2}$-norm loss (note the additional denominator of 2 for $w$ and $h$), which is consistent with the loss commonly used in horizontal detection. This also partly proves the correctness of using Wasserstein distance as the regression loss. See appendix for the detailed proof~\cite{Wasserstein2010djalil} of Eq.~\ref{eq:wd3}. 

\subsection{Gaussian Wasserstein Distance Regression Loss}

%Following the nonlinear transformation protocol widely used in existing IoU based regression loss to make the loss more smooth, the GWD term $d$ is wrapped by a nonlinear function $f$ which is finally converted into a 
Note that GWD alone can be sensitive to large errors. We perform a nonlinear transformation $f$ and then convert GWD into an affinity measure $\frac{1}{\tau+f(d^2)}$ similar to IoU between two bounding boxes. Then we follow the standard IoU based loss form in detection literature~\cite{rezatofighi2019generalized,zheng2020distance}, as written by:
%Note GWD Wcan be rather sensitive to large errors, hence we do not directly use GWD as the final regression loss, but first perform a nonlinear transformation and then convert it into a similarity variable. Finally, we use the form of IoU loss to construct the regression loss, which is given by:
\begin{equation}
\small
    \begin{aligned}
    L_{gwd} = 1-\frac{1}{\tau+f(d^2)}, \quad  \tau \geq 1
    \end{aligned}
	\label{eq:Lgwd}
\end{equation}
where $f(\cdot)$ denotes a non-linear function to transform the Wasserstein distance $d^2$ to make the loss more smooth and expressive. The hyperparameter $\tau$ modulates the entire loss.

Fig.~\ref{fig:iou_smooth-l1-1} plots the function curve under different different combinations of $f(\cdot)$ and $\tau$. Compared with the smooth L1 loss, the curve of Eq.~\ref{eq:Lgwd} is more consistent with the IoU loss curve. Furthermore, we can find in Fig.~\ref{fig:iou_smooth-l1-3} that GWD still can measure the distance between two non-overlapping bounding boxes (IoU=0), which is exactly the problem that GIoU and DIoU try to solve in horizontal detection. However, they cannot be applied for rotating detection.

Obviously, GWD has met the first two requirements in terms of consistency and differentiability with IoU loss. To analyze \textbf{Requirement 3}, we first give basic properties of Eq.~\ref{eq:wd1}:

\textbf{Property 1:} $\mathbf{\Sigma}^{1/2}(w,h,\theta)=\mathbf{\Sigma}^{1/2}(h,w,\theta-\frac{\pi}{2})$;

\textbf{Property 2:} $\mathbf{\Sigma}^{1/2}(w,h,\theta)=\mathbf{\Sigma}^{1/2}(w,h,\theta-\pi)$;

\textbf{Property 3:} $\mathbf{\Sigma}^{1/2}(w,h,\theta)\approx\mathbf{\Sigma}^{1/2}(w,h,\theta - \frac{\pi}{2})$, if $w\approx h$.

From the two bounding box definitions recall that the conversion between two definitions is, the two sides are exchanged and the angle difference is $90^\circ$. Many methods are designated inherently according to the choice of definition in advance to solve some problems, such as $D_{le}$ for EoE and $D_{oc}$ for square-like problem. It is interesting to note that according to \textbf{Property 1}, definition $D_{oc}$ and $D_{le}$ are equivalent for the GWD-based loss, which makes our method free from the choice of box definitions. This does not mean that the final performance of the two definition methods will be the same. Different factors such as angle definition and angle regression range will still cause differences in model learning, but the GWD-based method does not need to bind a certain definition method to solve the problem.

GWD can also help resolve the boundary discontinuity and square-like problem. The prediction box and ground truth in \textbf{way1} of \textbf{Case 1} in Fig.~\ref{fig:problems} satisfy the following relation: $x_{p}=x_{gt},y_{p}=y_{gt},w_{p}=h_{gt},h_{p}=w_{gt},\theta_{p}=\theta_{gt}-\frac{\pi}{2}$. According to \textbf{Property 1}, the Gaussian  distribution corresponding to these two boxes are the same (in the sense of same mean $\mathbf{m}$ and covariance $\mathbf{\Sigma}$), so it naturally eliminates the ambiguity in box representation. Similarly, according to \textbf{Properties 2-3}, the ground truth and prediction box in \textbf{way1} of \textbf{Case 1} and \textbf{Case 3} in Fig.~\ref{fig:problems} are also the same or nearly the same (note the approximate equal symbol for $w \approx h$ for square-like boxes) Gaussian distribution. Through the above analysis, we know GWD meets \textbf{Requirement 3}.

Overall, GWD is a unified solution to all the requirements and its advantages in rotating detection can be summarized:

%including inconsistency between metric and loss, boundary discontinuity and square-like problem. Specifically, GWD has the following advantages:
i) GWD makes the two bounding box definition methods equivalent, which enables our method to achieve significant improvement regardless how the bounding box is defined.

ii) GWD is a differentiable IoU loss approximation for rotating bounding box, which maintains a high consistency with the detection metric. GWD can also measure the distance between non-overlapping rotating bounding boxes and has properties similar to GIoU and DIoU for the horizontal case.
%so that GWD-based detector can achieve a huge performance improvement in high-precision detection.

iii) GWD inherently avoids the interference of boundary discontinuity and square-like problem, so that the model can learn in more diverse forms of regression, eliminate the inconsistency of regression under boundary and non-boundary positions, and reduce the learning cost.

\subsection{Overall Loss Function Design}
In line with~\cite{yang2020arbitrary,yang2020dense,yang2021r3det}, we use the one-stage detector RetinaNet~\cite{lin2017focal} as the baseline. Rotated rectangle is represented by five parameters ($x,y,w,h,\theta$). In our experiments we mainly follow $D_{oc}$, and the regression equation is as follows:
\begin{equation}
\small
	\begin{aligned}
	t_{x}&=(x-x_{a})/w_{a}, t_{y}=(y-y_{a})/h_{a} \\
	t_{w}&=\log(w/w_{a}), t_{h}=\log(h/h_{a}), t_{\theta}=\theta-\theta_{a}\\
	t_{x}^{*}&=(x_{}^{*}-x_{a})/w_{a}, t_{y}^{*}=(y_{}^{*}-y_{a})/h_{a} \\
	t_{w}^{*}&=\log(w_{}^{*}/w_{a}), t_{h}^{*}=\log(h_{}^{*}/h_{a}), t_{\theta}^{*}=\theta_{}^{*}-\theta_{a}
	\label{eq:regression}
	\end{aligned}
\end{equation}
where $x,y,w,h,\theta$ denote the box's center coordinates, width, height and angle, respectively. Variables $x, x_{a}, x^{*}$ are for the ground-truth box, anchor box, and predicted box, respectively (likewise for $y,w,h,\theta$). The multi-task loss is:
\begin{equation}
\small
	\begin{aligned}
	L = \frac{\lambda_{1}}{N}\sum_{n=1}^{N}obj_{n}\cdot L_{gwd}(b_{n},gt_{n}) + \frac{\lambda_{2}}{N}\sum_{n=1}^{N}L_{cls}(p_{n},t_{n})
	\label{eq:multitask_loss}
	\end{aligned}
\end{equation}
where $N$ indicates the number of anchors, $obj_{n}$ is a binary value ($obj_{n}=1$ for foreground and $obj_{n}=0$ for background, no regression for background). $b_{n}$ denotes the $n$-th predicted bounding box, $gt_{n}$ is the $n$-th target ground-truth. $t_{n}$ represents the label of $n$-th object, $p_{n}$ is the $n$-th probability distribution of various classes calculated by sigmoid function. The hyper-parameter $\lambda_{1}$, $\lambda_{2}$ control the trade-off and are set to $\{1,2\}$ by default. The classification loss $L_{cls}$ is set as the focal loss \cite{lin2017focal}.

%-------------------------------------------------------------------------

\begin{table}[tb!]
 %   \vskip 0.15in
    \begin{center}
    \begin{small}
    \begin{sc}
    \resizebox{0.43\textwidth}{!}{
        \begin{tabular}{c|cccc|c}
        \toprule
        % \diagbox{$f(\cdot)$}{Eq. \ref{eq:Lgwd}}{$\tau$} & 1 & 2 & 3 & 5 & $L_{gwd}=f(d^2)$ \\
        $1-\frac{1}{\left(\tau+f(d^2)\right)}$ & $\tau=1$ & $\tau=2$ & $\tau=3$ & $\tau=5$ & $d^2$ \\
        \hline
        $f(\cdot)=sqrt$ & 68.56 & \textbf{68.93} & 68.37 & 67.77 & \multirow{2}{*}{49.11} \\
        $f(\cdot)=\log$ & 67.87 & 68.09 & 67.48 & 66.49 & \\
        \bottomrule
        \end{tabular}}
    \end{sc}
    \end{small}
    \end{center}
    \caption{Ablation test of GWD-based regression loss form and hyperparameter on DOTA. The based detector is RetinaNet.}\vspace{-5pt}
    \label{tab:func_tau}
  %  \vskip -0.2in
\end{table}

\begin{table}[tb!]
   % \vskip 0.15in
    \begin{center}
    \begin{small}
    \begin{sc}
    \resizebox{0.48\textwidth}{!}{
        \begin{tabular}{c|c|c|c|c|c}
        \toprule
        Method & Box Def. & Reg. Loss & Dataset & Data Aug. & mAP$_{50}$ \\
        \hline
        \multirow{8}{*}{RetinaNet} & $D_{oc}$ & Smooth L1 & \multirow{2}{*}{HRSC2016} & \multirow{4}{*}{R+F+G} & 84.28 \\
        & $D_{oc}$ & GWD & & & \textbf{85.56 \color{red}{(+1.28)}}\\
        \cline{2-4} \cline{6-6}
        & $D_{oc}$ & Smooth L1 & \multirow{2}{*}{UCAS-AOD} & & 94.56 \\
        & $D_{oc}$ & GWD & & & \textbf{95.44 \color{red}{(+0.88)}}\\
        \cline{2-6}
        & $D_{oc}$ & Smooth L1 & \multirow{6}{*}{DOTA} & \multirow{6}{*}{F} & 65.73\\
        & $D_{oc}$ & GWD & & & \textbf{68.93 \color{red}{(+3.20)}}\\
        \cline{2-3} \cline{6-6}
        & $D_{le}$ & Smooth L1 &  &  & 64.17\\
        & $D_{le}$ & GWD & & & \textbf{66.31 \color{red}{(+2.14)}}\\
        \cline{1-3} \cline{6-6}
        \multirow{2}{*}{R$^3$Det} & $D_{oc}$  & Smooth L1 & & & 70.66 \\
        & $D_{oc}$ & GWD & & & \textbf{71.56 \color{red}{(+0.90)}}\\
        \bottomrule
        \end{tabular}}
    \end{sc}
    \end{small}
    \end{center}
    \caption{Ablation study for GWD on three datasets. `R', `F' and `G' indicate random rotation, flipping, and graying, respectively.}\vspace{-5pt}
    \label{tab:ablation_rs}
  %  \vskip -0.2in
\end{table}

\begin{table}[tb!]
   % \vskip 0.15in
    \begin{center}
    \begin{small}
    \begin{sc}
    \resizebox{0.48\textwidth}{!}{
        \begin{tabular}{c|c|c|c|c|c|c}
        \toprule
        Method & Reg. Loss & Dataset & Data Aug. & Recall & Precision & Hmean \\
        \hline
        \multirow{6}{*}{RetinaNet} & Smooth L1 & \multirow{2}{*}{MLT} & \multirow{4}{*}{F} & 37.88 & 67.07 & 48.42 \\
        & GWD & & & 44.01 & 71.83 & \textbf{54.58 \color{red}{(+6.16)}}\\
        \cline{2-3} \cline{5-7}
        & Smooth L1 & \multirow{8}{*}{ICDAR2015} & & 71.55 & 68.10 & 69.78\\
        & GWD & & & 73.95 & 74.64 & \textbf{74.29 \color{red}{(+4.51)}}\\
        \cline{2-2} \cline{4-7}
        & Smooth L1 & & \multirow{2}{*}{R+F} & 69.43 & 81.15 & 74.83\\
        & GWD & & & 72.17 & 80.59 & \textbf{76.15 \color{red}{(+1.32)}}\\
        \cline{1-2} \cline{4-7}
        \multirow{4}{*}{R$^3$Det} & Smooth L1 & & \multirow{2}{*}{F} & 69.09 & 80.30 & 74.28\\
        & GWD & & & 70.00 & 82.15 & \textbf{75.59 \color{red}{(+1.31)}} \\
        \cline{2-2} \cline{4-7}
        & Smooth L1 & & \multirow{2}{*}{R+F} & 71.69 & 79.80 & 75.53 \\
        & GWD & & & 73.95 & 80.50 & \textbf{77.09 \color{red}{(+1.56)}}\\    
        \bottomrule
        \end{tabular}}
    \end{sc}
    \end{small}
    \end{center}
    \caption{Ablation study for GWD on two scene text datasets.}\vspace{-5pt}
    \label{tab:ablation_st}
   % \vskip -0.2in
\end{table}

\begin{table*}[tb!]
  %  \vskip 0.15in
    \begin{center}
    \begin{small}
    \begin{sc}
    \resizebox{1.0\textwidth}{!}{
			\begin{tabular}{c|c|c|c|ccccc|ccccccccccccccc|c}
				\toprule
				ID & Moethod & Backbone & Sched. & DA & MS & MSC & SWA & ME & PL &  BD &  BR &  GTF &  SV &  LV &  SH &  TC &  BC &  ST & SBF &  RA &  HA &  SP &  HC &  mAP$_{50}$\\
				\hline
                \textbf{\#1} & \multirow{10}{*}{RetinaNet-GWD} & \multirow{2}{*}{R-50} & \multirow{2}{*}{20} & & & & & &88.49 & 77.88 & 44.07 & 66.08 & 71.92 & 62.56 & 77.94 & 89.75 & 81.43 & 79.64 & 52.30 & 63.52 & 60.25 & 66.51 & 51.63 & 68.93\\
                \textbf{\#2} & & & & & & & $\checkmark$ & & 88.60 & 78.59 & 44.10 & 67.24 & 70.77 & 62.54 & 79.78 & 88.86 & 81.92 & 80.46 & 57.44 & 64.02 & 62.64 & 66.52 & 55.29 & 69.92\\
                \cline{3-25}
                \textbf{\#3} & & \multirow{7}{*}{R-152} & \multirow{2}{*}{40} & $\checkmark$ & & & & & 89.06 & 83.48 & 49.84 & 65.34 & 74.64 & 67.63 & 82.39 & 88.39 & 84.19 & 84.80 & 63.74 & 61.32 & 66.47 & 70.94 & 67.52 & 73.32 \\
                \textbf{\#4} & & & & $\checkmark$ & $\checkmark$ & & & & 87.47 & 83.77 & 52.30 & 68.24 & 73.24 & 65.14 & 80.18 & 89.63 & 84.39 & 85.53 & 65.79 & 66.02 & 69.57 & 72.21 & 69.79 & 74.22 \\
                \cline{4-25}
                \textbf{\#5} & & & \multirow{5}{*}{60} & $\checkmark$ & & & & & 88.88 &	80.47 &	52.94 &	63.85 & 76.95 & 	70.28 &	83.56 &	88.54 &	83.51 &	84.94 &	61.24 & 65.13 & 65.45 & 71.69 & 73.90 & 74.09\\
                \textbf{\#6} & & & & $\checkmark$ & $\checkmark$ & & & & 87.12 &	81.64 &	54.79 &	68.74 &	76.17 &	68.39 & 83.93 & 89.06 & 84.51 & 85.99 & 63.33 & 66.68 & 72.60 & 70.63 & 74.17 & 75.18\\
                \textbf{\#7} & & & & $\checkmark$ & $\checkmark$ & $\checkmark$ & & & 86.14 & 81.59 & 55.33 & 75.57 & 74.20 & 67.34 &	81.75 &	87.48 &	82.80 &	85.46 &	69.47 &	67.20 &	70.97 & 70.91 & 74.07 & 75.35\\
                \textbf{\#8} & & & & $\checkmark$ & $\checkmark$ & & $\checkmark$ & & 87.63 & 84.32 & 54.83 & 69.99 & 76.17 & 70.12 &	83.13 &	88.96 &	83.19 &	86.06 &	67.72 &	66.17 &	73.47 &	74.57 &	72.80 &	75.94\\
                \textbf{\#9} & & & & $\checkmark$ & $\checkmark$ & $\checkmark$ & $\checkmark$ & & 86.96 & 83.88 & 54.36 & 77.53 & 74.41 &	68.48 &	80.34 &	86.62 &	83.41 &	85.55 &	73.47 &	67.77 &	72.57 &	75.76 &	73.40 &	76.30\\
                \cline{3-25}
                \textbf{\#10} &  & -- & -- & $\checkmark$ & $\checkmark$ & $\checkmark$ & $\checkmark$ & $\checkmark$ & \textbf{\color{blue}{89.06}} & \textbf{\color{blue}{84.32}} & \textbf{\color{blue}{55.33}} & \textbf{\color{blue}{77.53}} & \textbf{\color{blue}{76.95}} & \textbf{\color{blue}{70.28}} & \textbf{\color{blue}{83.95}} & \textbf{\color{blue}{89.75}} & \textbf{\color{blue}{84.51}} & \textbf{\color{blue}{86.06}} & \textbf{\color{blue}{73.47}} & \textbf{\color{blue}{67.77}} & \textbf{\color{blue}{72.60}} & \textbf{\color{blue}{75.76}} & \textbf{\color{blue}{74.17}} & \textbf{\color{blue}{77.43}} \\
                \cline{2-25}
                \textbf{\#11} & \multirow{24}{*}{R$^3$Det-GWD} & \multirow{5}{*}{R-101} & \multirow{8}{*}{30} & $\checkmark$ & & & & & 89.59 & 81.18 & 52.89 & 70.37 & 77.73 & 82.42 & 86.99 & 89.31 & 83.06 & 85.97 & 64.07 & 65.14 & 68.05 & 70.95 & 58.45 & 75.08\\
                \textbf{\#12} & & & & $\checkmark$ &$\checkmark$ & & & & 89.64 & 81.70 & 52.52 & 72.96 & 76.02 & 82.60 & 87.17 & 89.57 & 81.25 & 86.09 & 62.24 & 65.74 & 68.05 & 74.96 & 64.38 & 75.66\\
                \textbf{\#13} & & & & $\checkmark$ & $\checkmark$ & & $\checkmark$ & & 89.66 & 82.11 & 52.74 & 71.64 & 75.95 & 83.09 & 86.97 & 89.28 & 85.04 & 86.17 & 65.52 & 63.29 & 72.18 & 74.88 & 63.17 & 76.11\\
                \textbf{\#14} & & & & $\checkmark$ & $\checkmark$ & $\checkmark$ & & & 89.56 & 81.23 &	53.38 &	79.38 &	75.12 &	82.14 &	86.86 &	88.87 &	81.21 &	86.28 &	65.36 &	65.06 &	72.88 &	73.04 &	62.97 &	76.22\\
                \textbf{\#15} & & & & $\checkmark$ & $\checkmark$ & $\checkmark$ & $\checkmark$ & & 89.33 & 80.86 & 53.28 & 78.29 & 75.40 & 82.69 & 87.09 & 89.35 & 82.64 & 86.41 & 69.85 & 64.71 & 74.19 & 76.18 & 59.85 & 76.67\\
                \cline{3-3}  \cline{5-25}
                \textbf{\#16} & & \multirow{3}{*}{R-152} & & $\checkmark$ & & & & & 89.51 &	82.68 &	51.92 & 69.51 &	78.97 &	83.38 &	87.53 & 89.67 &	85.65 &	86.17 &	63.90 &	67.44 &	68.27 &	76.43 &	64.22 &	76.35\\
                \textbf{\#17} & & & & $\checkmark$ & $\checkmark$ & & & & 89.55 & 82.28 &	52.39 &	68.30 &	77.86 &	83.40 &	87.48 &	89.56 &	84.27 &	86.14 &	65.38 &	63.25 &	71.33 &	72.36 &	69.21 &	76.18\\
                \textbf{\#18} & & & & $\checkmark$ & $\checkmark$ & $\checkmark$ & & & 89.62 & 82.27 & 52.35 & 77.30 & 76.95 & 82.53 & 87.20 & 89.08 & 84.58 & 86.21 & 65.21 & 64.46 & 74.99 & 76.30 & 65.19 & 76.95 \\
                \cline{3-25}
                \textbf{\#19} & & \multirow{5}{*}{R-18} & \multirow{5}{*}{40} & $\checkmark$ & & & & & 86.63 & 80.12 & 51.98 & 49.67 & 75.73 & 77.54 & 86.10 & 90.05 & 83.22 & 82.31 & 56.05 & 58.86 & 63.30 & 69.06 & 55.07 & 71.05\\
                \textbf{\#20} & & & & $\checkmark$ & $\checkmark$ & & & & 87.88 & 81.73 & 51.76 & 69.21 & 73.78 & 77.78 & 86.46 & 90.05 & 84.47 & 84.33 & 59.82 & 59.74 & 66.54 & 69.15 & 60.42 & 73.54\\
                \textbf{\#21} & & & & $\checkmark$ & $\checkmark$ & & $\checkmark$ & & 88.94 & 84.10 & 53.04 & 67.78 & 75.29 & 79.21 & 86.89 & 89.90 & 86.43 & 84.30 & 63.22 & 59.96 & 67.16 & 70.55 & 64.39 & 74.74\\
                \textbf{\#22} & & & & $\checkmark$ & $\checkmark$ & $\checkmark$ & & & 87.27 & 82.59 & 51.90 & 76.58 & 72.74 & 77.04 & 85.59 & 89.18 & 83.91 & 84.81 & 63.34 & 59.46 & 66.41 & 69.79 & 59.03 & 73.98\\
                \textbf{\#23} & & & & $\checkmark$ & $\checkmark$ & $\checkmark$ & $\checkmark$ & & 88.38 & 84.75 & 52.63 & 77.35 & 74.29 & 78.53 & 86.32 & 89.12 & 85.73 & 85.13 & 67.84 & 59.48 & 66.88 & 71.59 & 62.58 & 75.37 \\
                \cline{3-25}
                \textbf{\#24} & & \multirow{5}{*}{R-50} & \multirow{10}{*}{60} & $\checkmark$ & & & & & 88.82 & 82.94 & 55.63 & 72.75 & 78.52 &	83.10 &	87.46 &	90.21 &	86.36 &	85.44 &	64.70 &	61.41 &	73.46 &	76.94 &	57.38 &	76.34\\
                \textbf{\#25} & & & & $\checkmark$ & $\checkmark$ & & & & 89.09 &	84.13 &	55.77 &	74.48 &	77.71 & 82.99 & 87.57 & 89.46 & 84.89 & 85.67 &	66.09 &	64.17 &	75.13 &	75.35 & 62.78 &	77.02\\
                \textbf{\#26} & & & & $\checkmark$ & $\checkmark$ & & $\checkmark$ & & 89.04 & 84.99 & 57.14 & 76.13 & 77.79 & 84.03 & 87.70 & 89.53 & 83.83 & 85.64 & 69.60 & 63.75 & 76.10 & 79.22 & 67.80 & 78.15\\
                \textbf{\#27} & & & & $\checkmark$ & $\checkmark$ & $\checkmark$ & & & 88.89 & 83.58 & 55.54 &	80.46 &	76.86 &	83.07 &	86.85 &	89.09 &	83.09 &	86.17 &	71.38 &	64.93 &	76.21 &	73.23 &	64.39 &	77.58\\
                \textbf{\#28} & & & & $\checkmark$ & $\checkmark$ & $\checkmark$ & $\checkmark$ & & 88.43 & 84.33 & 56.91 & 82.19 & 76.69 & 83.23 & 86.78 & 88.90 & 83.93 & 85.73 & 72.07 & 65.67 & 76.76 & 78.37 & 65.31 & 78.35\\
                \cline{3-3}  \cline{5-25}
                \textbf{\#29} & & \multirow{5}{*}{R-152} & & $\checkmark$ & & & & & 88.74 & 82.63 & 54.88 & 70.11 & 78.87 & 84.59 & 87.37 & 89.81 & 84.79 & 86.47 & 66.58 & 64.11 & 75.31 & 78.43 & 70.87 & 77.57\\
                \textbf{\#30} & & & & $\checkmark$ & $\checkmark$ & & & & 89.59 & 84.19 & 56.53 & 75.69 & 77.67 & 84.48 & 87.52 & 90.05 & 84.29 & 86.85 & 68.61 & 64.73 & 76.59 & 77.92 & 71.88 & 78.44\\
                \textbf{\#31} & & & & $\checkmark$ & $\checkmark$ & & $\checkmark$ & & 89.59 & 82.96 & 58.83 & 75.04 & 77.63 & 84.83 & 87.31 & 89.89 & 86.54 & 86.82 & 69.45 & 65.94 & 76.55 & 77.50 & 74.92 & 78.92\\
                \textbf{\#32} & & & & $\checkmark$ & $\checkmark$ & $\checkmark$ & & & 88.99 & 82.26 & 56.62 & 81.40 & 77.04 & 83.90 & 86.56 & 88.97 & 83.63 & 86.48 & 70.45 & 65.58 & 76.41 & 77.30 & 69.21 & 78.32\\
                \textbf{\#33} & & & & $\checkmark$ & $\checkmark$ & $\checkmark$ & $\checkmark$ & & 89.28 & 83.70 & 59.26 & 79.85 & 76.42 & 83.87 & 86.53 & 89.06 & 85.53 & 86.50 & 73.04 & 67.56 & 76.92 & 77.09 & 71.58 & 79.08\\
                \cline{3-25}
                \textbf{\#34} &  & -- & -- & $\checkmark$ & $\checkmark$ & $\checkmark$ & $\checkmark$ & $\checkmark$ & \textbf{\color{blue}{89.66}} & \textbf{\color{blue}{84.99}} & \textbf{\color{blue}{59.26}} & \textbf{\color{blue}{82.19}} & \textbf{\color{blue}{78.97}} & \textbf{\color{blue}{84.83}} & \textbf{\color{blue}{87.70}} & \textbf{\color{blue}{90.21}} & \textbf{\color{blue}{86.54}} & \textbf{\color{blue}{86.85}} & \textbf{\color{blue}{73.04}} & \textbf{\color{blue}{67.56}} & \textbf{\color{blue}{76.92}} & \textbf{\color{blue}{79.22}} & \textbf{\color{blue}{74.92}} & \textbf{\color{blue}{80.19}}\\
                \cline{2-25}
                \textbf{\#35} & -- & -- & -- & $\checkmark$ & $\checkmark$ & $\checkmark$ & $\checkmark$ & $\checkmark$ & \textbf{\color{red}{89.66}} & \textbf{\color{red}{84.99}} & \textbf{\color{red}{59.26}} & \textbf{\color{red}{82.19}} & \textbf{\color{red}{78.97}} & \textbf{\color{red}{84.83}} & \textbf{\color{red}{87.70}} & \textbf{\color{red}{90.21}} & \textbf{\color{red}{86.54}} & \textbf{\color{red}{86.85}} & \textbf{\color{red}{73.47}} & \textbf{\color{red}{67.77}} & \textbf{\color{red}{76.92}} & \textbf{\color{red}{79.22}} & \textbf{\color{red}{74.92}} & \textbf{\color{red}{80.23}}\\
				\bottomrule
		\end{tabular}} 
    \end{sc}
    \end{small}
    \end{center}
    \caption{Ablation experiment of training strategies and tricks. R-101 denotes ResNet-101 (likewise for R-18, R-50, R-152). MS, MSC, SWA, and ME represent data augmentation, multi-scale training and testing, stochastic weights averaging, multi-scale image cropping, and model ensemble, respectively. }
    \label{tab:SOTA_}
    %\vskip -0.2in
\end{table*}

\begin{table}[tb!]
  %  \vskip 0.15in
    \begin{center}
    \begin{small}
    \begin{sc}
    \resizebox{0.48\textwidth}{!}{
        \begin{tabular}{c|c|cccc|c}
        \toprule
        Method & Reg. Loss & AP$_{50}$ & AP$_{60}$ & AP$_{75}$ & AP$_{85}$ & AP$_{50:95}$ \\
        \hline
        \multirow{2}{*}{RetinaNet} &
        Smooth L1 & 84.28 & 74.74 & 48.42 & 12.56 & 47.76 \\
        & GWD & \textbf{85.56} & \textbf{84.04} & \textbf{60.31} & \textbf{17.14} & \textbf{52.89 \color{red}{+(5.13)}} \\
        \hline
        \multirow{2}{*}{R$^3$Det} &
        Smooth L1 & 88.52 & 79.01 & 43.42 & 4.58 & 46.18 \\
        & GWD & \textbf{89.43} & \textbf{88.89} & \textbf{65.88} & \textbf{15.02} & \textbf{56.07 \color{red}{+(9.89)}} \\
        \bottomrule
        \end{tabular}}
    \end{sc}
    \end{small}
    \end{center}
    \caption{High-precision detection experiment on HRSC206 data set. The image resolution is 512, and data augmentation is used.}\vspace{-5pt}
    \label{tab:ablation_ap}
   % \vskip -0.2in
\end{table}

\begin{table*}[tb!]
    \caption{Comparison between different solutions for inconsistency between metric and loss (IML), boundary discontinuity (BD) and square-like problem (SLP) on DOTA dataset. The $\checkmark$ indicates that the method has corresponding problem. $^\dagger$ and $^\ddagger$ represent the large aspect ratio object and the square-like object, respectively. The bold \textbf{\color{red}{red}} and \textbf{\color{blue}{blue}} fonts indicate the top two performances respectively.}\vspace{-10pt}
    \label{tab:ablation_study}
   % \vskip 0.15in
    \begin{center}
    \begin{small}
    \begin{sc}
    \resizebox{1.0\textwidth}{!}{
        \begin{tabular}{c|c|c|c|cc|c|ccccc|cc|cc|ccc}
        \toprule
        \multirow{2}{*}{Base Detector} & \multirow{2}{*}{Method} & \multirow{2}{*}{Box Def.} & \multirow{2}{*}{IML} & \multicolumn{2}{c|}{BD} & \multirow{2}{*}{SLP} & \multicolumn{9}{c|}{tranval/test} & \multicolumn{3}{c}{train/val} \\
        \cline{5-6} \cline{8-19}
        & & & & EoE & PoA & & BR$^\dagger$ & SV$^\dagger$ & LV$^\dagger$ & SH$^\dagger$ & HA$^\dagger$ & ST$^\ddagger$ & RA$^\ddagger$ & 7-mAP$_{50}$ & mAP$_{50}$ & mAP$_{50}$ & mAP$_{75}$ & mAP$_{50:95}$\\
        \hline
        \multirow{7}{*}{RetinaNet} & - & $D_{oc}$ & $\checkmark$ & $\checkmark$ & $\checkmark$ & $\times$ & 42.17 & 65.93 & 51.11 & 72.61 & 53.24 & 78.38 & 62.00 & 60.78 & 65.73 & 64.70 & 32.31 & 34.50 \\
        & - & $D_{le}$ & $\checkmark$ & $\checkmark$ & $\checkmark$ & $\checkmark$ & 38.31 & 60.48 & 49.77 & 68.29 & 51.28 & 78.60 & 60.02 & 58.11 & 64.17 & 62.21 & 26.06 & 31.49\\
        & IoU-Smooth L1 Loss & $D_{oc}$ & $\checkmark$ & $\times$ & $\times$ & $\times$ & \textbf{\color{red}{44.32}} & 63.03 & 51.25 & 72.78 & \textbf{\color{blue}{56.21}} & 77.98 & \textbf{\color{blue}{63.22}} & 61.26 & 66.99 & 64.61 & 34.17 & 36.23\\
        & Modulated Loss & $D_{oc}$ & $\checkmark$ & $\times$ & $\times$ & $\times$ & 42.92 & 67.92 & 52.91 & 72.67 & 53.64 & \textbf{\color{red}{80.22}} & 58.21 & 61.21 & 66.05 & 63.50 & 33.32 & 34.61\\
        % & Modulated Loss & quad. & 30 & $\checkmark$ & $\times$ & $\times$ & $\times$ & 43.31 & \textbf{\color{blue}{68.49}} & \textbf{\color{blue}{61.02}} & \textbf{\color{red}{79.62}} & \textbf{\color{blue}{58.43}} & 76.72 & \textbf{\color{blue}{63.50}} & \textbf{\color{blue}{64.44}} & 67.27 & 62.28 & \textbf{\color{blue}{37.82}} & 36.23\\
        & CSL & $D_{le}$ & $\checkmark$ & $\times$ & $\times$ & $\checkmark$ & 42.25 & \textbf{\color{blue}{68.28}} & 54.51 & 72.85 & 53.10 & 75.59 & 58.99 & 60.80 & 67.38 & 64.40 & 32.58 & 35.04 \\
        & DCL (BCL) & $D_{le}$ & $\checkmark$ & $\times$ & $\times$ & $\times$ & 41.40 & 65.82 & \textbf{\color{blue}{56.27}} & \textbf{\color{blue}{73.80}} & 54.30 & 79.02 & 60.25 & \textbf{\color{blue}{61.55}} & \textbf{\color{blue}{67.39}} & \textbf{\color{red}{65.93}} & \textbf{\color{blue}{35.66}} & \textbf{\color{blue}{36.71}} \\
        & GWD & $D_{oc}$ & $\times$ & $\times$ & $\times$ & $\times$ & \textbf{\color{blue}{44.07}} & \textbf{\color{red}{71.92}} & \textbf{\color{red}{62.56}} & \textbf{\color{red}{77.94}} & \textbf{\color{red}{60.25}} & \textbf{\color{blue}{79.64}} & \textbf{\color{red}{63.52}} & \textbf{\color{red}{65.70}} & \textbf{\color{red}{68.93}} & \textbf{\color{blue}{65.44}} & \textbf{\color{red}{38.68}} & \textbf{\color{red}{38.71}} \\
        \cline{1-19}
        \multirow{3}{*}{R$^3$Det} & - & $D_{oc}$ & $\checkmark$ & $\checkmark$ & $\checkmark$ & $\times$ & 44.15 & \textbf{\color{blue}{75.09}} & 72.88 & \textbf{\color{blue}{86.04}} & 56.49 & 82.53 & 61.01 & 68.31 & 70.66 & 67.18 & \textbf{\color{blue}{38.41}} & \textbf{\color{blue}{38.46}} \\
        & DCL (BCL) & $D_{le}$ & $\checkmark$ & $\times$ & $\times$ & $\times$ & \textbf{\color{red}{46.84}} & 74.87 & \textbf{\color{blue}{74.96}} & 85.70 & \textbf{\color{blue}{57.72}} & \textbf{\color{red}{84.06}} & \textbf{\color{red}{63.77}} & \textbf{\color{blue}{69.70}} & \textbf{\color{blue}{71.21}} & \textbf{\color{blue}{67.45}} & 35.44 & 37.54\\
        & GWD & $D_{oc}$ & $\times$ & $\times$ & $\times$ & $\times$ & \textbf{\color{blue}{46.73}} & \textbf{\color{red}{75.84}} & \textbf{\color{red}{78.00}} & \textbf{\color{red}{86.71}} & \textbf{\color{red}{62.69}} & \textbf{\color{blue}{83.09}} & \textbf{\color{blue}{61.12}} & \textbf{\color{red}{70.60}} & \textbf{\color{red}{71.56}} & \textbf{\color{red}{69.28}} & \textbf{\color{red}{43.35}} & \textbf{\color{red}{41.56}}\\
        \bottomrule
        \end{tabular}}
    \end{sc}
    \end{small}
    \end{center}
    \vskip -0.15in
\end{table*}

\begin{table*}[tb!]
        \begin{center}
        \begin{small}
        \begin{sc}
		\resizebox{1.0\textwidth}{!}{
			\begin{tabular}{l|lccccccccccccccccc|c}
				\toprule
				& Method & Backbone & MS &  PL &  BD &  BR &  GTF &  SV &  LV &  SH &  TC &  BC &  ST &  SBF &  RA &  HA &  SP &  HC &  mAP$_{50}$\\
				\midrule
				\multirow{27}{*}{\rotatebox{90}{\shortstack{Two-stage methods}}}
				& FR-O \cite{xia2018dota} & R-101 & & 79.09 & 69.12 & 17.17 & 63.49 & 34.20 & 37.16 & 36.20 & 89.19 & 69.60 & 58.96 & 49.4 & 52.52 & 46.69 & 44.80 & 46.30 & 52.93 \\
				&ICN \cite{azimi2018towards} & R-101 & $\checkmark$ & 81.40 & 74.30 & 47.70 & 70.30 & 64.90 & 67.80 & 70.00 & 90.80 & 79.10 & 78.20 & 53.60 & 62.90 & 67.00 & 64.20 & 50.20 & 68.20 \\
				& KARNET \cite{tang2020rotating} & R-50 & & 89.33 & 83.55 & 44.79 & 71.61 & 63.05 & 67.06 & 69.53 & 90.47 & 79.46 & 77.84 & 51.04 & 60.97 & 65.38 & 69.46 & 49.53 & 68.87\\
				&RADet \cite{li2020radet} & RX-101 & & 79.45 & 76.99 & 48.05 & 65.83 & 65.46 & 74.40 & 68.86 & 89.70 & 78.14 & 74.97 & 49.92 & 64.63 & 66.14 & 71.58 & 62.16 & 69.09 \\
				&RoI-Trans. \cite{ding2018learning} & R-101 & $\checkmark$ & 88.64 & 78.52 & 43.44 & 75.92 & 68.81 & 73.68 & 83.59 & 90.74 & 77.27 & 81.46 & 58.39 & 53.54 & 62.83 & 58.93 & 47.67 & 69.56 \\
				&CAD-Net \cite{zhang2019cad} & R-101 & & 87.8 & 82.4 & 49.4 & 73.5 & 71.1 & 63.5 & 76.7 & 90.9 & 79.2 & 73.3 & 48.4 & 60.9 & 62.0 & 67.0 & 62.2 & 69.9 \\
				& AOOD \cite{zou2020arbitrary} & DPN-92 & $\checkmark$  & 89.99 & 81.25 & 44.50 & 73.20 & 68.90 & 60.33 & 66.86 & 90.89 & 80.99 & 86.23 & 64.98 & 63.88 & 65.24 & 68.36 & 62.13 & 71.18 \\
				& Cascade-FF \cite{hou2020cascade} & R-152 & & 89.9 & 80.4 & 51.7 & 77.4 & 68.2 & 75.2 & 75.6 & 90.8 & 78.8 & 84.4 & 62.3 & 64.6 & 57.7 & 69.4 & 50.1 & 71.8\\
				& SCRDet \cite{yang2019scrdet} & R-101 & $\checkmark$ & 89.98 & 80.65 & 52.09 & 68.36 & 68.36 & 60.32 & 72.41 & 90.85 & \textbf{\color{blue}{87.94}} & 86.86 & 65.02 & 66.68 & 66.25 & 68.24 & 65.21 & 72.61\\
				& SARD \cite{wang2019sard} & R-101 & & 89.93 & 84.11 & 54.19 & 72.04 & 68.41 & 61.18 & 66.00 & 90.82 & 87.79 & 86.59 & 65.65 & 64.04 & 66.68 & 68.84 & 68.03 & 72.95 \\
				& GLS-Net \cite{li2020object} & R-101 & & 88.65 & 77.40 & 51.20 & 71.03 & 73.30 & 72.16 & 84.68 & 90.87 & 80.43 & 85.38 & 58.33 & 62.27 & 67.58 & 70.69 & 60.42 & 72.96 \\
				&FADet \cite{li2019feature} & R-101 & $\checkmark$ & 90.21 & 79.58 & 45.49 & 76.41 & 73.18 & 68.27 & 79.56 & 90.83 & 83.40 & 84.68 & 53.40 & 65.42 & 74.17 & 69.69 & 64.86 & 73.28\\
				& MFIAR-Net \cite{yang2020multi} & R-152 & $\checkmark$ & 89.62 & 84.03 & 52.41 & 70.30 & 70.13 & 67.64 & 77.81 & 90.85 & 85.40 & 86.22 & 63.21 & 64.14 & 68.31 & 70.21 & 62.11 & 73.49 \\
				&Gliding Vertex \cite{xu2020gliding} & R-101 & & 89.64 & 85.00 & 52.26 & 77.34 & 73.01 & 73.14 & 86.82 & 90.74 & 79.02 & 86.81 & 59.55 & 70.91 & 72.94 & 70.86 & 57.32 & 75.02 \\
				& SAR \cite{lu2020sar} & R-152 & & 89.67 & 79.78 & 54.17 & 68.29 & 71.70 & 77.90 & 84.63 & 90.91 & \textbf{\color{red}{88.22}} & 87.07 & 60.49 & 66.95 & 75.13 & 75.28 & 64.29 & 75.28\\
				&Mask OBB \cite{wang2019mask} & RX-101 & $\checkmark$ & 89.56 & 85.95 & 54.21 & 72.90 & 76.52 & 74.16 & 85.63 & 89.85 & 83.81 & 86.48 & 54.89 & 69.64 & 73.94 & 69.06 & 63.32 & 75.33 \\
				&FFA \cite{fu2020rotation} & R-101 & $\checkmark$ & \textbf{\color{blue}{90.1}} & 82.7 & 54.2 & 75.2 & 71.0 & 79.9 & 83.5 & 90.7 & 83.9 & 84.6 & 61.2 & 68.0 & 70.7 & 76.0 & 63.7 & 75.7 \\
				&APE \cite{zhu2020adaptive} & RX-101 & & 89.96 & 83.62 & 53.42 & 76.03 & 74.01 & 77.16 & 79.45 & 90.83 & 87.15 & 84.51 & 67.72 & 60.33 & 74.61 & 71.84 & 65.55 & 75.75 \\
				& F$^3$-Net \cite{ye2020f3} & R-152 & $\checkmark$ & 88.89 & 78.48 & 54.62 & 74.43 & 72.80 & 77.52 & 87.54 & 90.78 & 87.64 & 85.63 & 63.80 & 64.53 & \textbf{\color{blue}{78.06}} & 72.36 & 63.19 & 76.02 \\
				&CenterMap \cite{wang2020learning} & R-101 & $\checkmark$ & 89.83 & 84.41 & 54.60 & 70.25 & 77.66 & 78.32 & 87.19 & 90.66 & 84.89 & 85.27 & 56.46 & 69.23 & 74.13 & 71.56 & 66.06 & 76.03\\
				&CSL \cite{yang2020arbitrary} & R-152 & $\checkmark$ & \textbf{\color{red}{90.25}} & 85.53 & 54.64 & 75.31 & 70.44 & 73.51 & 77.62 & 90.84 & 86.15 & 86.69 & 69.60 & 68.04 & 73.83 & 71.10 & 68.93 & 76.17\\
				& MRDet \cite{qin2020mrdet} & R-101 & & 89.49 & 84.29 & 55.40 & 66.68 & 76.27 & 82.13 & 87.86 & 90.81 & 86.92 & 85.00 & 52.34 & 65.98 & 76.22 & 76.78 & 67.49 & 76.24\\
				&RSDet-II \cite{qian2021learning} & R-152 & $\checkmark$ & 89.93& 84.45 & 53.77 & 74.35 & 71.52 & 78.31 & 78.12 & \textbf{\color{red}{91.14}} & 87.35 & 86.93 & 65.64 & 65.17 & 75.35 & 79.74 & 63.31 & 76.34 \\
				& OPLD \cite{song2020learning} & R-101 & $\checkmark$ & 89.37 & \textbf{\color{blue}{85.82}} & 54.10 & 79.58 & 75.00 & 75.13 & 86.92 & 90.88 & 86.42 & 86.62 & 62.46 & 68.41 & 73.98 & 68.11 & 63.69 & 76.43 \\
				& SCRDet++ \cite{yang2022scrdet++} & R-101 & $\checkmark$ & 90.05 & 84.39 & 55.44 & 73.99 & 77.54 & 71.11 & 86.05 & 90.67 & 87.32 & 87.08 & 69.62 & 68.90 & 73.74 & 71.29 & 65.08 & 76.81\\
				& HSP \cite{xu2020hierarchical} & R-101 & $\checkmark$ & 90.39 & \textbf{\color{red}{86.23}} & 56.12 & \textbf{\color{blue}{80.59}} & 77.52 & 73.26 & 83.78 & 90.80 & 87.19 & 85.67 & 69.08 & \textbf{\color{red}{72.02}} & 76.98 & 72.50 & 67.96 & 78.01\\
				& FR-Est \cite{fu2020point} & R-101-DCN & $\checkmark$ & 89.78 & 85.21 & 55.40 & 77.70 & \textbf{\color{red}{80.26}} & \textbf{\color{blue}{83.78}} & 87.59 & 90.81 & 87.66 & 86.93 & 65.60 & 68.74 & 71.64 & \textbf{\color{red}{79.99}} & 66.20 & 78.49\\
				% & SCRDet++ \cite{yang2022scrdet++} & R-152 & $\checkmark$ & 90.06 & 84.37 & 56.76 & 75.82 & 79.33 & 74.50 & 86.51 & 90.77 & 88.51 & 86.84 & 69.60 & 69.05 & 76.15 & 79.92 & 76.61 & 78.99\\
				\hline
				\multirow{20}{*}{\rotatebox{90}{\shortstack{Single-stage methods}}} & IENet \cite{lin2019ienet} & R-101 & $\checkmark$ & 80.20 & 64.54 & 39.82 & 32.07 & 49.71 & 65.01 & 52.58 & 81.45 & 44.66 & 78.51 & 46.54 & 56.73 & 64.40 & 64.24 & 36.75 & 57.14 \\
				& TOSO \cite{feng2020toso} & R-101 & $\checkmark$ & 80.17 & 65.59 & 39.82 & 39.95 & 49.71 & 65.01 & 53.58 & 81.45 & 44.66 & 78.51 & 48.85 & 56.73 & 64.40 & 64.24 & 36.75 & 57.92\\
				&PIoU \cite{chen2020piou} & DLA-34 & & 80.9 & 69.7 & 24.1 & 60.2 & 38.3 & 64.4 & 64.8 & \textbf{\color{blue}{90.9}} & 77.2 & 70.4 & 46.5 & 37.1 & 57.1 & 61.9 & 64.0 & 60.5 \\
				& Axis Learning \cite{xiao2020axis} & R-101 & & 79.53 & 77.15 & 38.59 & 61.15 & 67.53 & 70.49 & 76.30 & 89.66 & 79.07 & 83.53 & 47.27 & 61.01 & 56.28 & 66.06 & 36.05 & 65.98\\
				& A$^2$S-Det \cite{xiao2021a2s} & R-101 & & 89.59 & 77.89 & 46.37 & 56.47 & 75.86 & 74.83 & 86.07 & 90.58 & 81.09 & 83.71 & 50.21 & 60.94 & 65.29 & 69.77 & 50.93 & 70.64 \\
				&O$^2$-DNet \cite{wei2020oriented} & H-104 & $\checkmark$ & 89.31 & 82.14 & 47.33 & 61.21 & 71.32 & 74.03 & 78.62 & 90.76 & 82.23 & 81.36 & 60.93 & 60.17 & 58.21 & 66.98 & 61.03 & 71.04 \\
				& P-RSDet \cite{zhou2020arbitrary} & R-101 & $\checkmark$ & 88.58 & 77.83 & 50.44 & 69.29 & 71.10 & 75.79 & 78.66 & 90.88 & 80.10 & 81.71 & 57.92 & 63.03 & 66.30 & 69.77 & 63.13 & 72.30 \\
				&BBAVectors \cite{yi2020oriented} & R-101 & $\checkmark$ & 88.35 & 79.96 & 50.69 & 62.18 & 78.43 & 78.98 & 87.94 & 90.85 & 83.58 & 84.35 & 54.13 & 60.24 & 65.22 & 64.28 & 55.70 & 72.32 \\
				& ROPDet \cite{yang2020ropdet} & R-101-DCN & $\checkmark$ & 90.01 & 82.82 & 54.47 & 69.65 & 69.23 & 70.78 & 75.78 & 90.84 & 86.13 & 84.76 & 66.52 & 63.71 & 67.13 & 68.38 & 46.09 & 72.42\\
				& HRP-Net \cite{he2020high} & HRNet-W48 & & 89.33 & 81.64 & 48.33 & 75.21 & 71.39 & 74.82 & 77.62 & 90.86 & 81.23 & 81.96 & 62.93 & 62.17 & 66.27 & 66.98 & 62.13 & 72.83\\
				&DRN \cite{pan2020dynamic} & H-104 & $\checkmark$ & 89.71 & 82.34 & 47.22 & 64.10 & 76.22 & 74.43 & 85.84 & 90.57 & 86.18 & 84.89 & 57.65 & 61.93 & 69.30 & 69.63 & 58.48 & 73.23 \\
				& CFC-Net \cite{ming2021cfc} & R-101 & $\checkmark$ & 89.08 & 80.41 & 52.41 & 70.02 & 76.28 & 78.11 & 87.21 & 90.89 & 84.47 & 85.64 & 60.51 & 61.52 & 67.82 & 68.02 & 50.09 & 73.50\\
				% & EFN \cite{liu2020efn} & U-Net & $\checkmark$ & 93.44 & 76.38 & 37.05 & 78.47 & 88.75 & 89.96 & 90.58 & 90.91 & 94.89 & 78.02 & 63.87 & 57.41 & 40.73 & 95.49 & 53.13 & 75.27\\
				& R$^4$Det \cite{sun2020r4} & R-152 & & 88.96 & 85.42 & 52.91 & 73.84 & 74.86 & 81.52 & 80.29 & 90.79 & 86.95 & 85.25 & 64.05 & 60.93 & 69.00 & 70.55 & 67.76 & 75.84 \\
				&R$^3$Det \cite{yang2021r3det} & R-152 & $\checkmark$ & 89.80 & 83.77 & 48.11 & 66.77 & 78.76 & 83.27 & 87.84 & 90.82 & 85.38 & 85.51 & 65.67 & 62.68 & 67.53 & 78.56 & \textbf{\color{blue}{72.62}} & 76.47\\
				& PolarDet \cite{zhao2020polardet} & R-101 & $\checkmark$ & 89.65 & 87.07 & 48.14 & 70.97 & 78.53 & 80.34 & 87.45 & 90.76 & 85.63 & 86.87 & 61.64 & 70.32 & 71.92 & 73.09 & 67.15 & 76.64 \\
				& S$^2$A-Net-DAL \cite{ming2020dynamic} & R-50 & $\checkmark$ & 89.69 & 83.11 & 55.03 & 71.00 & 78.30 & 81.90 & \textbf{\color{blue}{88.46}} & 90.89 & 84.97 & \textbf{\color{blue}{87.46}} & 64.41 & 65.65 & 76.86 & 72.09 & 64.35 & 76.95 \\
				& R$^3$Det-DCL \cite{yang2020dense} & R-152 & $\checkmark$ & 89.26 & 83.60 & 53.54 & 72.76 & 79.04 & 82.56 & 87.31 & 90.67 & 86.59 & 86.98 & 67.49 & 66.88 & 73.29 & 70.56 & 69.99 & 77.37 \\
				& RDD \cite{zhong2020single} & R-101 & $\checkmark$ & 89.15 & 83.92 & 52.51 & 73.06 & 77.81 & 79.00 & 87.08 & 90.62 & 86.72 & 87.15 & 63.96 & \textbf{\color{blue}{70.29}} & 76.98 & 75.79 & 72.15 & 77.75 \\
				% \cline{2-20}
				% & S$^2$A-Net & R-50 & $\checkmark$ & 88.89 & 83.60 & 57.74 & 81.95 & 79.94 & 83.19 & 89.11 & 90.78 & 84.87 & 87.81 & 70.30 & 68.25 & 78.30 & 77.01 & 69.58 & 79.42 \\
				& S$^2$A-Net \cite{han2020align} & R-101 & $\checkmark$ & 89.28 & 84.11 & \textbf{\color{blue}{56.95}} & 79.21 & \textbf{\color{blue}{80.18}} & 82.93 & \textbf{\color{red}{89.21}} & 90.86 & 84.66 & \textbf{\color{red}{87.61}} & \textbf{\color{blue}{71.66}} & 68.23 & \textbf{\color{red}{78.58}} & 78.20 & 65.55 & \textbf{\color{blue}{79.15}}\\
				\cline{2-20}
				& GWD (Ours) & R-152 & $\checkmark$ & 89.66 & 84.99 & \textbf{\color{red}{59.26}} & \textbf{\color{red}{82.19}} & 78.97 & \textbf{\color{red}{84.83}} & 87.70 & 90.21 & 86.54 & 86.85 & \textbf{\color{red}{73.47}} & 67.77 & 76.92 & \textbf{\color{blue}{79.22}} & \textbf{\color{red}{74.92}} & \textbf{\color{red}{80.23}} \\
				
				\bottomrule
		\end{tabular}}
		\end{sc}
        \end{small}
        \end{center}
        \caption{AP on different objects and mAP on DOTA. R-101 denotes ResNet-101 (likewise for R-50, R-152), RX-101 and H-104 stands for ResNeXt101~\cite{xie2017aggregated} and Hourglass-104~\cite{newell2016stacked}. Other backbone include DPN-92 \cite{chen2017dual}, DLA-34 \cite{yu2018deep}, DCN \cite{dai2017deformable}, HRNet-W48 \cite{wang2020deep}, U-Net \cite{ronneberger2015u}. MS indicates that multi-scale training or testing is used.}
        \label{tab:DOTA_SOTA}
\end{table*}

\begin{table}[tb!]
        \begin{center}
        \begin{small}
        \begin{sc}
		\resizebox{0.48\textwidth}{!}{
			\begin{tabular}{lccc}
				\toprule
				
				Method & Backbone & mAP$_{50}$ (07) & mAP$_{50}$ (12)\\
				
				\hline
				RC1 \& RC2 \cite{liu2017high} & VGG16 & 75.7 & -- \\
				Axis Learning \cite{xiao2020axis} & R-101 & 78.15 & --\\
				TOSO \cite{feng2020toso} & R-101 & 79.29 & -- \\ 
				R$^2$PN \cite{zhang2018toward}  & VGG16 & 79.6 & --  \\
				RRD \cite{liao2018rotation} & VGG16  & 84.3 & --  \\
				RoI-Trans. \cite{ding2018learning} & R-101 & 86.20 & -- \\
				RSDet \cite{qian2021learning} & R-50 & 86.50 & --\\
				DRN \cite{pan2020dynamic} & H-104 & -- & 92.70 \\
				CenterMap \cite{wang2020learning} & R-50 & -- & 92.8 \\
				SBD \cite{liu2019omnidirectional} & R-50 & -- & 93.70 \\
				Gliding Vertex \cite{xu2020gliding} & R-101 & 88.20 & -- \\
				OPLD  \cite{song2020learning} & R-101 & 88.44 & --\\
				BBAVectors \cite{yi2020oriented} & R-101 & 88.6 & -- \\
				S$^2$A-Net \cite{han2020align} & R-101 & \textbf{\color{red}{90.17}} & 95.01 \\
				R$^3$Det \cite{yang2021r3det} & R-101 & 89.26 & 96.01 \\
                R$^3$Det-DCL \cite{yang2020dense} & R-101 & 89.46 & \textbf{\color{blue}{96.41}}\\
                FPN-CSL \cite{yang2020arbitrary} & R-101 & 89.62 & 96.10 \\
                DAL \cite{ming2020dynamic} & R-101 & 89.77 & -- \\
                \hline
                R$^3$Det-GWD (Ours) & R-101 & \textbf{\color{blue}{89.85}} & \textbf{\color{red}{97.37}}\\
				\bottomrule
		\end{tabular}}
		\end{sc}
        \end{small}
        \end{center}
        \caption{Detection accuracy on HRSC2016.}
        \label{tab:HRSC2016_SOTA}
\end{table}

\section{Experiments}
\label{sec:experiment}
We use Tensorflow \cite{abadi2016tensorflow} for implementation on a server with Tesla V100 and 32G memory.
\subsection{Datasets and Implementation Details}
\textbf{DOTA} \cite{xia2018dota} is comprised of 2,806 large aerial images from different sensors and platforms. Objects in DOTA exhibit a wide variety of scales, orientations, and shapes. These images are then annotated by experts using 15 object categories. 
The short names for categories are defined as (abbreviation-full name): PL-Plane, BD-Baseball diamond, BR-Bridge, GTF-Ground field track, SV-Small vehicle, LV-Large vehicle, SH-Ship, TC-Tennis court, BC-Basketball court, ST-Storage tank, SBF-Soccer-ball field, RA-Roundabout, HA-Harbor, SP-Swimming pool, and HC-Helicopter. 
The fully annotated DOTA benchmark contains 188,282 instances, each of which is labeled by an arbitrary quadrilateral. 
Half of the original images are randomly selected as the training set, 1/6 as the validation set, and 1/3 as the testing set. We divide the images into $ 600 \times 600 $ subimages with an overlap of 150 pixels and scale it to $ 800 \times 800 $. With all these processes, we obtain about 20,000 training and 7,000 validation patches.

\textbf{UCAS-AOD} \cite{zhu2015orientation} contains 1,510 aerial images of about $ 659 \times 1,280 $ pixels, with 2 categories of 14,596  instances. In line with~\cite{azimi2018towards,xia2018dota}, we sample 1,110 images for training and 400 for testing. 

\textbf{HRSC2016} \cite{liu2017high} contains images from two scenarios including ships on sea and ships close inshore. 
%All images are collected from six famous harbors. 
The training, validation and test set include 436, 181 and 444 images, respectively.

\textbf{ICDAR2015} \cite{karatzas2015icdar} is commonly used for oriented scene text detection and spotting. This dataset includes 1,000 training images and 500 testing images. 

\textbf{ICDAR 2017 MLT} \cite{nayef2017icdar2017} is a multi-lingual text dataset, which includes 7,200 training images, 1,800 validation images and 9,000 testing images. The dataset is composed of complete scene images in 9 languages, and text regions in this dataset can be in arbitrary orientations, being more diverse and challenging.

Experiments are initialized by ResNet50 \cite{he2016deep} by default unless otherwise specified. We perform experiments on three aerial benchmarks and two scene text benchmarks to verify the generality of our techniques. Weight decay and momentum are set 0.0001 and 0.9, respectively. We employ MomentumOptimizer over 8 GPUs with a total of 8 images per mini-batch (1 image per GPU).
All the used datasets are trained by 20 epochs in total, and learning rate is reduced tenfold at 12 epochs and 16 epochs, respectively. The initial learning rates for RetinaNet is 5e-4. The number of image iterations per epoch for DOTA, UCAS-AOD, HRSC2016, ICDAR2015, and MLT are 54k, 5k, 10k, 10k and 10k respectively, and increase exponentially if data augmentation and multi-scale training are used. 

\subsection{Ablation Study}

\textbf{Ablation test of GWD-based regression loss form and hyperparameter:} 
Tab. \ref{tab:func_tau} compares two different forms of GWD-based loss. The performance of directly using GWD ($d^2$) as the regression loss is extremely poor, only 49.11\%, due to its rapid growth trend. In other words, the regression loss $d^2$ is too sensitive to large errors. In contrast, Eq. \ref{eq:Lgwd} achieves a significant improvement by fitting IoU loss. Eq. \ref{eq:Lgwd} introduces two new hyperparameters, the non-linear function $f(\cdot)$ to transform the Wasserstein distance, and the constant $\tau$ to modulate the entire loss. From Tab. \ref{tab:func_tau}, the overall performance of using $sqrt$ outperforms that using $\log$, about 0.98$\pm$0.3\% higher. 
%Indeed, in Fig.~\ref{fig:iou_smooth-l1}, the $sqrt$-based GWD curve is closer to the IoU loss curve than the $\log$-based GWD curve. 
For $f(\cdot)=sqrt$ with $\tau=2$, the model achieves the best performance, about 68.93\%. All the subsequent experiments follow this setting for hyperparameters unless otherwise specified.

\textbf{Ablation test with different rotating box definitions:} 
As mentioned above, definition $D_{oc}$ and $D_{le}$ are equivalent for the GWD-based loss according to \textbf{Property 1}, which makes our method free from the choice of box definitions. This does not mean that the final performance of the two definition methods will be the same, but that the GWD-based method does not need to bind a certain definition method to solve the boundary discontinuity or square-like problem. Tab. \ref{tab:ablation_rs} compares the performance of RetinaNet under different regression loss on DOTA, and both rotating box definitions: $D_{le}$ and $D_{oc}$ are tested. For the smooth L1 loss, the accuracy of $D_{le}$-based method is 1.56\% lower than $D_{le}$-based, at 64.17\% and 65.73\%, respectively. GWD-based method does not need to be coupled with a certain definition to solve boundary discontinuity or square-like problem, it has increased by 2.14\% and 3.20\% under above two definitions. 

\textbf{Ablation test across datasets and detectors:} We use two detectors on five datasets to verify the effectiveness of GWD. When RetinaNet is used as the base detector in Tab. \ref{tab:ablation_rs}, the GWD-based detector is improved by 1.28\%, 0.88\%, 3.20\%, 2.14\% under three different aerial image datasets of HRSC206, UCAS-AOD and DOTA, respectively. Note that to increase the reliability of the results from small dataset, the experiments of the first two datasets have involved additional data augmentation, including random graying and random rotation. The rotation detector R$^3$Net~\cite{yang2021r3det} achieves the state-of-the-art performance on large-scale DOTA. It can be seen that GWD further improves the performance by 0.90\%. Tab. \ref{tab:ablation_st} also gives ablation test on two scene text datasets. There are a large number of objects in the boundary position in scene text, so the GWD-based RetinaNet has obtained a notable gain -- increased by 6.16\% and 4.51\% on the MLT and ICDAR2015 datasets, respectively. Even with the use of data augmentation or a stronger detector R$^3$Det, GWD can still obtain a stable gain, with an improvement range from 1.31\% to 1.56\%.

\subsection{Training Strategies and Tricks}
In order to further improve the performance of the model on DOTA, we verified many commonly used training strategies and tricks, including backbone, training schedule, data augmentation (DA), multi-scale training and testing (MS), stochastic weights averaging (SWA) \cite{izmailov2018averaging, zhang2020swa}, multi-scale image cropping (MSC), model ensemble (ME), as shown in Tab. \ref{tab:SOTA_}.

\textbf{Backbone:} 
Under the conditions of different detectors (RetinaNet and R$^3$Det), different training schedules (experimental groups \{\textbf{\#11},\textbf{\#16}\}, \{\textbf{\#24},\textbf{\#29}\}), and different tricks (experimental groups \{\textbf{\#26},\textbf{\#31}\}, \{\textbf{\#28},\textbf{\#33}\}), large backbone can bring stable performance improvement.

\textbf{Multi-scale training and testing:}
Multi-scale training and testing is an effective means to improve the performance of aerial images with various object scales. In this paper, training and testing scale set to [450, 500, 640, 700, 800, 900, 1,000, 1,100, 1,200]. Experimental groups \{\textbf{\#3},\textbf{\#4}\}, \{\textbf{\#5},\textbf{\#6}\} and \{\textbf{\#11},\textbf{\#12}\} show the its effectiveness, increased by 0.9\%, 1.09\%, and 0.58\%, respectively.

\textbf{Training schedule:}
When data augmentation and multi-scale training are added, it is necessary to appropriately lengthen the training time. From the experimental groups \{\textbf{\#3},\textbf{\#5}\} and \{\textbf{\#16},\textbf{\#29}\}, we can find that the performance respectively increases by 0.77\% and 1.22\% when the training schedule is increased from 40  or 30 epochs to 60 epochs.

\textbf{Stochastic weights averaging (SWA):}
SWA technique has been proven to be an effective tool for improving object detection. In the light of \cite{zhang2020swa}, we train our detector for an extra 12 epochs using cyclical learning rates and then average these 12 checkpoints as the final detection model. It can be seen from experimental groups \{\textbf{\#1}, \textbf{\#2}\}, \{\textbf{\#20}, \textbf{\#21}\} and \{\textbf{\#25}, \textbf{\#26}\} in Tab. \ref{tab:SOTA_} that we get 0.99\%,  1.20\% and 1.13\% improvement on the challenging DOTA benchmark.

\textbf{Multi-scale image cropping:}
Large-scene object detection often requires image sliding window cropping before training. During testing, sliding window cropping testing is required before the results are merged. Two adjacent sub-images often have an overlapping area to ensure that the truncated object can appear in a certain sub-image completely. The cropping size needs to be moderate, too large is not conducive to the detection of small objects, and too small will cause large objects to be truncated with high probability. Multi-scale cropping is an effective detection technique that is beneficial to objects of various scales. In this paper, our multi-scale crop size and corresponding overlap size are [600, 800, 1,024, 1,300, 1,600] and [150, 200, 300, 300, 400], respectively. According to experimental groups \{\textbf{\#6}, \textbf{\#7}\} and \{\textbf{\#30}, \textbf{\#32}\}, the large object categories (e.g. GTF and SBF) that are often truncated have been significantly improved. Take group \{\textbf{\#6}, \textbf{\#7}\} as an example, GTF and SBF increased by 6.43\% and 6.14\%, respectively.

\subsection{Further Comparison}
\textbf{High precision detection:}
The advantage of aligning detection metric and loss is that a higher precision prediction box can be learned. Object with large aspect ratios are more sensitive to detection accuracy, so we conduct high-precision detection experiments on the ship dataset HRSC2016. It can be seen in Tab. \ref{tab:ablation_ap} that our GWD-based detector exhibits clear advantages under high IoU thresholds. Taking AP$_{75}$ as an example, GWD has achieved improvement by 11.89\% and 22.46\% on the two detectors, respectively. We also compares the peer techniques, mainly including IoU-Smooth L1 Loss~\cite{yang2019scrdet}, CSL~\cite{yang2020arbitrary}, and DCL~\cite{yang2020dense} on DOTA validation set. As shown on the right of Tab. \ref{tab:ablation_study}, the GWD-based method achieves the highest performance on mAP$_{75}$ and mAP$_{50:95}$, at 38.68\% and 38.71\%.

\textbf{Comparison of techniques to solve the regression issues:} For the three issues of inconsistency between metric and loss, boundary discontinuity and square-like problem, Tab. \ref{tab:ablation_study} compares the five peer techniques, including IoU-Smooth L1 Loss, Modulated loss, CSL, and DCL on DOTA test set. For fairness, these methods are all implemented on the same baseline method, and are trained and tested under the same environment and hyperparameters.

In particular, we detail the accuracy of the seven categories, including large aspect ratio (e.g. BR, SV, LV, SH, HA) and square-like object (e.g. ST, RD), which contain many corner cases in the dataset. These categories are assumed can better reflect the real-world challenges and advantages of our method. Many methods that solve the boundary discontinuity have achieved significant improvements in the large aspect ratio object category, and the methods that take into account the square-like problem perform well in the square-like object, such as GWD, DCL and Modulated loss.

However, there is rarely a unified method to solve all problems, and most methods are proposed for part of problems. Among them, the most comprehensive method is IoU-Smooth L1 Loss. However, the gradient direction of IoU-Smooth L1 Loss is still dominated by smooth L1 loss, so the metric and loss cannot be regarded as truly consistent. Besides, IoU-Smooth L1 Loss needs to determine whether the prediction box is within the defined range before calculating IoU at the boundary position, Otherwise, it needs to convert to the same definition as ground truth. In contrast, due to the three unique properties of GWD, it need to make additional judgments to elegantly solve all problems. From Tab. \ref{tab:ablation_study}, GWD outperforms on most categories. For the seven listed categories (7-mAP) and overall performance (mAP), GWD-based methods are also the best. Fig.~\ref{fig:compare_vis} visualizes the comparison between Smooth L1 loss-based and GWD-based detector. 

\subsection{Comprehensive Overall Comparison}

\textbf{Results on DOTA:} Due to the complexity of the aerial image and the large number of small, cluttered and rotated objects, DOTA is a very challenging dataset. We compare the proposed approach
with other state-of-the-art methods on DOTA, as shown in Tab. \ref{tab:DOTA_SOTA}. As far as I know, this is the most comprehensive statistical comparison of methods on the DOTA dataset. Since different methods use different image resolution, network structure, training strategies and various tricks, we cannot make absolutely fair comparisons. In terms of overall performance, our method has achieved the best performance so far, at around 80.23\%.

\textbf{Results on HRSC2016:} The HRSC2016 contains lots of large aspect ratio ship instances with arbitrary orientation, which poses a huge challenge to the positioning accuracy of the detector. Experimental results at Tab. \ref{tab:HRSC2016_SOTA} shows that our model achieves state-of-the-art performances, about 89.85\% and 97.37\% in term of 2007 and 2012 evaluation metric.

%-------------------------------------------------------------------------
\section{Conclusion}
\label{sec:conclusion}
This paper has presented a Gaussian Wasserstain distance based loss to model the deviation between two rotating bounding boxes for object detection. The designated loss directly aligns with the detection accuracy and the model can be efficiently learned via back-propagation. More importantly, thanks to its three unique properties, GWD can also elegantly solve the boundary discontinuity and square-like problem regardless how the bounding box is defined. Experimental results on extensive public benchmarks show the state-of-the-art performance of our detector.

%-------------------------------------------------------------------------
\section*{Appendix}
\subsection{Proof of $d:=\mathbf{W}(\mathcal{N}(\mathbf{m}_{1},\mathbf{\Sigma}_{1});\mathcal{N}(\mathbf{m}_{2},\mathbf{\Sigma}_{2}))$}
The entire proof process refers to this blog~\cite{Wasserstein2010djalil}.

The Wasserstein coupling distance $\mathbf{W}$ between two probability measures $\mu$ and $\nu$ on $\mathbb{R}^n$ expressed as follows:
\begin{equation}
    \begin{aligned}
        \mathbf{W}(\mu;\nu):= \inf \mathbb{E}(\lVert \mathbf{X}-\mathbf{Y}\rVert_{2}^{2})^{1/2}
    \end{aligned}
	\label{eq:wd2_}
\end{equation}
where the infimum runs over all random vectors $(\mathbf{X},\mathbf{Y})$ of $\mathbb{R}^n\times \mathbb{R}^n$ with $\mathbf{X} \sim \mu$ and $\mathbf{Y} \sim \nu$. It turns out that we have the following formula for $d:=\mathbf{W}(\mathcal{N}(\mathbf{m}_{1},\mathbf{\Sigma}_{1});\mathcal{N}(\mathbf{m}_{2},\mathbf{\Sigma}_{2}))$:
\begin{equation}
    \begin{aligned}
        d^{2}=\lVert \mathbf{m}_{1}-\mathbf{m}_{2}\rVert_{2}^{2}+\mathbf{Tr}\left(\mathbf{\Sigma}_{1}+\mathbf{\Sigma}_{2}-2(\mathbf{\Sigma}_{1}^{1/2}\mathbf{\Sigma}_{2}\mathbf{\Sigma}_{1}^{1/2})^{1/2}\right)
    \end{aligned}
	\label{eq:wd3_}
\end{equation}
This formula interested several works~\cite{givens1984class,olkin1982distance,knott1984optimal,dowson1982frechet}. Note in particular we have:
\begin{equation}
    \begin{aligned}
        \mathbf{Tr}\left((\mathbf{\Sigma}_{1}^{1/2}\mathbf{\Sigma}_{2}\mathbf{\Sigma}_{1}^{1/2})^{1/2}\right)=\mathbf{Tr}\left((\mathbf{\Sigma}_{2}^{1/2}\mathbf{\Sigma}_{1}\mathbf{\Sigma}_{2}^{1/2})^{1/2}\right)
    \end{aligned}
	\label{eq:wd4_}
\end{equation}

In the commutative case  $\mathbf{\Sigma}_{1}\mathbf{\Sigma}_{2}=\mathbf{\Sigma}_{2}\mathbf{\Sigma}_{1}$, Eq.~\ref{eq:wd3_} becomes:
\begin{equation}
    \begin{aligned}
        d^{2}=&\lVert \mathbf{m}_{1}-\mathbf{m}_{2}\rVert_{2}^{2}+\lVert \mathbf{\Sigma}_{1}^{1/2}-\mathbf{\Sigma}_{2}^{1/2} \rVert_{F}^{2}\\
        =&(x_{1}-x_{2})^2+(y_{1}-y_{2})^2+\frac{(w_{1}-w_{2})^2+(h_{1}-h_{2})^2}{4}\\
        =&l_{2}\text{-norm}\left(\left[x_1,y_1,\frac{w_1}{2},\frac{h_1}{2}\right]^\top,\left[x_2,y_2,\frac{w_2}{2},\frac{h_2}{2}\right]^\top\right)
    \end{aligned}
	\label{eq:wd5_}
\end{equation}
where $\|\|_F$ is the Frobenius norm. Note that both boxes are horizontal at this time, and Eq.~\ref{eq:wd5_} is approximately equivalent to the $l_{2}$-norm loss (note the additional denominator of 2 for $w$ and $h$), which is consistent with the loss commonly used in horizontal detection. This also partly proves the correctness of using Wasserstein distance as the regression loss. 

To prove Eq.~\ref{eq:wd3_}, one can first reduce to the centered case $\mathbf{m}_{1}=\mathbf{m}_{2}=\mathbf{0}$. Next, if $(\mathbf{X},\mathbf{Y})$ is a random vector (Gaussian or not) of $\mathbb{R}^n\times \mathbb{R}^n$ with covariance matrix
\begin{equation}
    \begin{aligned}
        \mathbf{\Gamma}=
        \left(                 
          \begin{array}{cc}   
            \mathbf{\Sigma}_{1} & \mathbf{C}\\  
            \mathbf{C}^{\top} & \mathbf{\Sigma}_{2}\\  
          \end{array}
        \right)
    \end{aligned}
	\label{eq:wd6_}
\end{equation}
then the quantity
\begin{equation}
    \begin{aligned}
        \mathbb{E}(\lVert \mathbf{X},\mathbf{Y}\rVert_{2}^{2})=\mathbf{Tr}(\mathbf{\Sigma}_{1}+\mathbf{\Sigma}_{2}-2\mathbf{C})
    \end{aligned}
	\label{eq:wd7_}
\end{equation}
depends only on $\mathbf{\Gamma}$. Also, when $\mu=\mathcal{N}(\mathbf{0},\mathbf{\Sigma}_{1})$ and $\nu=\mathcal{N}(\mathbf{0},\mathbf{\Sigma}_{2})$, one can restrict the infimum which defines $W$ to run over Gaussian laws $\mathcal{N}(\mathbf{0},\mathbf{\Gamma})$ on $\mathbb{R}^n\times \mathbb{R}^n$ with covariance matrix $\mathbf{\Gamma}$ structured as above. The sole constrain on $\mathbf{C}$ is the Schur complement constraint:
\begin{equation}
    \begin{aligned}
        \mathbf{\Sigma}_{1}-\mathbf{C}\mathbf{\Sigma}_{2}^{-1}\mathbf{C}^{\top}\succeq 0
    \end{aligned}
	\label{eq:wd8_}
\end{equation}
The minimization of the function
\begin{equation}
    \begin{aligned}
        \mathbf{C}\rightarrowtail -2\mathbf{Tr}(\mathbf{C})
    \end{aligned}
	\label{eq:wd9_}
\end{equation}

\begin{figure}[!tb]
	\begin{center}
		\includegraphics[width=0.95\linewidth]{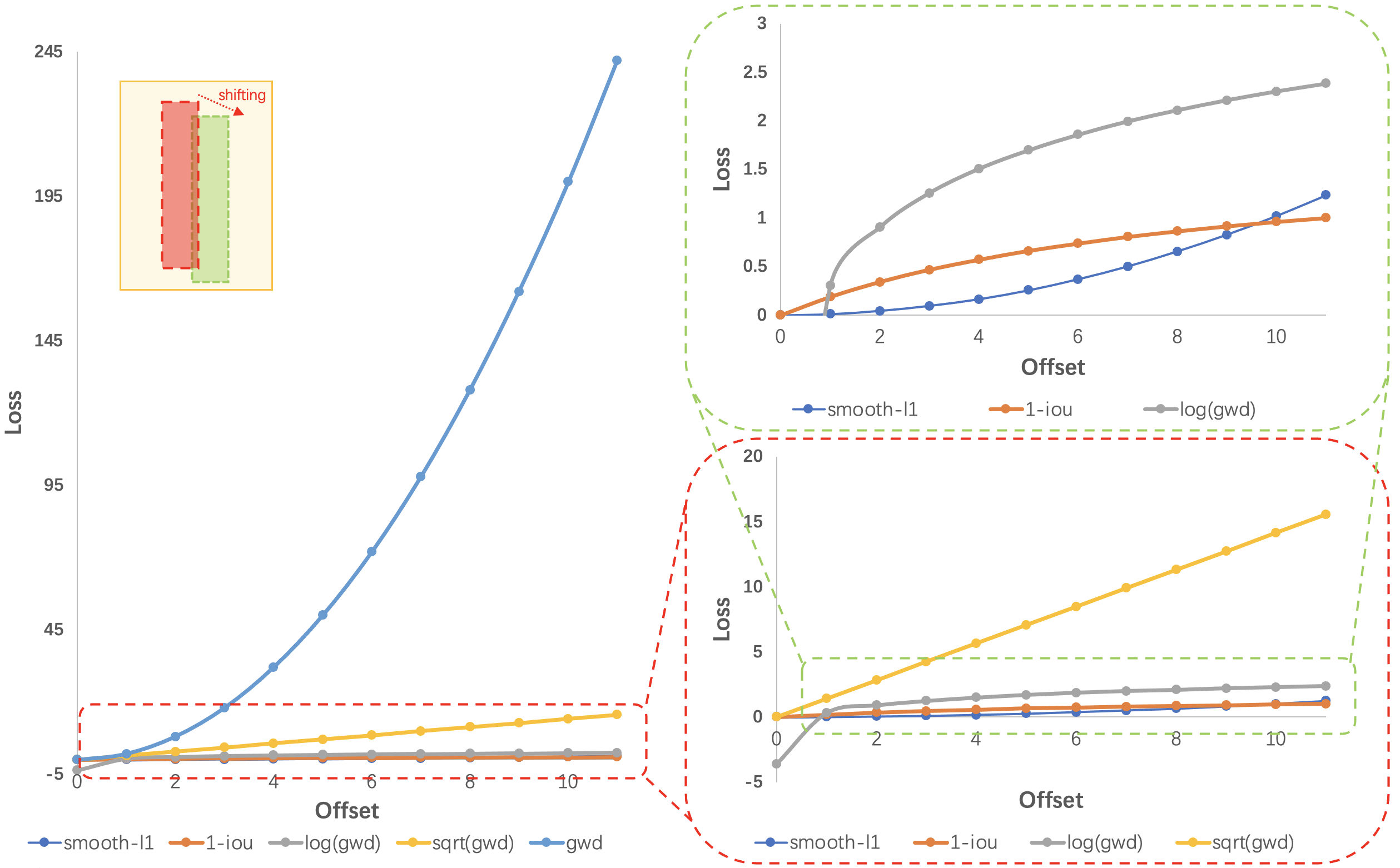}
	\end{center}
	\caption{Different forms of GWD-based regression loss curve.}
	\label{fig:loss_form}
\end{figure}

under the constraint above leads to Eq.~\ref{eq:wd3_}. A detailed proof is given by \cite{givens1984class}. Alternatively, one may find an optimal transportation map as \cite{knott1984optimal}. It turns out that $\mathcal{N}(\mathbf{m}_{2},\mathbf{\Sigma}_{2})$ is the image law of $\mathcal{N}(\mathbf{m}_{1},\mathbf{\Sigma}_{1})$ with the linear map
\begin{equation}
    \begin{aligned}
        \mathbf{x}\rightarrowtail \mathbf{m}_{2}+\mathbf{A}(\mathbf{x}\mathbf{m}_{1})
    \end{aligned}
	\label{eq:wd10_}
\end{equation}
where
\begin{equation}
    \begin{aligned}
        \mathbf{A}= \mathbf{\Sigma}_{1}^{-1/2}(\mathbf{\Sigma}_{1}^{1/2} \mathbf{\Sigma}_{2}\mathbf{\Sigma}_{1}^{1/2})^{1/2}\mathbf{\Sigma}_{1}^{-1/2}=\mathbf{A}^{\top}
    \end{aligned}
	\label{eq:wd11_}
\end{equation}
To check that this maps $\mathcal{N}(\mathbf{m}_{1},\mathbf{\Sigma}_{1})$ to $\mathcal{N}(\mathbf{m}_{2},\mathbf{\Sigma}_{2})$, say in the case $\mathbf{m}_{1}=\mathbf{m}_{2}=\mathbf{0}$ for simplicity, one may define the random column vectors $\mathbf{X}\sim\mathcal{N}(\mathbf{m}_{1},\mathbf{\Sigma}_{1})$ and $\mathbf{Y}=\mathbf{A}\mathbf{X}$ and write
\begin{equation}
    \begin{aligned}
        \mathbb{E}(\mathbf{Y}\mathbf{Y}^{\top})=&\mathbf{A}\mathbb{E}(\mathbf{X}\mathbf{X}^{\top})\mathbf{A}^{\top}\\
        =&\mathbf{\Sigma}_{1}^{1/2}(\mathbf{\Sigma}_{1}^{1/2}\mathbf{\Sigma}_{2}\mathbf{\Sigma}_{1}^{1/2})^{1/2}(\mathbf{\Sigma}_{1}^{1/2}\mathbf{\Sigma}_{2}\mathbf{\Sigma}_{1}^{1/2})^{1/2}\mathbf{\Sigma}_{1}^{1/2}\\
        =&\mathbf{\Sigma}_{2}
    \end{aligned}
	\label{eq:wd12_}
\end{equation}
To check that the map is optimal, one may use,
\begin{equation}
    \begin{aligned}
        \mathbb{E}(\lVert \mathbf{X}-\mathbf{Y} \rVert_{2}^{2})=&\mathbb{E}(\lVert \mathbf{X} \rVert_{2}^{2})+\mathbb{E}(\lVert \mathbf{Y} \rVert_{2}^{2})-2\mathbb{E}(<\mathbf{X},\mathbf{Y}>)\\
        =&\mathbf{Tr}(\mathbf{\Sigma}_{1})+\mathbf{Tr}(\mathbf{\Sigma}_{2})-2\mathbb{E}(<\mathbf{X},\mathbf{A}\mathbf{X}>)\\
        =&\mathbf{Tr}(\mathbf{\Sigma}_{1})+\mathbf{Tr}(\mathbf{\Sigma}_{2})-2\mathbf{Tr}(\mathbf{\Sigma}_{1}\mathbf{A})
    \end{aligned}
	\label{eq:wd13_}
\end{equation}
and observe that by the cyclic property of the trace,
\begin{equation}
    \begin{aligned}
        \mathbf{Tr}(\mathbf{\Sigma}_{1}\mathbf{A})=\mathbf{Tr}((\mathbf{\Sigma}_{1}^{1/2}\mathbf{\Sigma}_{2}\mathbf{\Sigma}_{1}^{1/2})^{1/2})
    \end{aligned}
	\label{eq:wd14_}
\end{equation}
The generalizations to elliptic families of distributions and to infinite dimensional Hilbert spaces is probably easy. Some more “geometric” properties of Gaussians with respect to such distances where studied more recently by \cite{takatsu2012cone} and \cite{takatsu2012cone}.

\begin{table}[tb!]
 %   \vskip 0.15in
    \begin{center}
    \begin{small}
    \begin{sc}
    \resizebox{0.48\textwidth}{!}{
        \begin{tabular}{c|cccc|c|c}
        \toprule
        % \diagbox{$f(\cdot)$}{Eq. \ref{eq:Lgwd}}{$\tau$} & 1 & 2 & 3 & 5 & $L_{gwd}=f(d^2)$ \\
        $1-\frac{1}{\left(\tau+f(d^2)\right)}$ & $\tau=1$ & $\tau=2$ & $\tau=3$ & $\tau=5$ & $f(d^2)$ & $d^2$ \\
        \hline
        $f(\cdot)=sqrt$ & 68.56 & \textbf{\color{blue}{68.93}} & 68.37 & 67.77 & 54.27 & \multirow{2}{*}{49.11} \\
        $f(\cdot)=\log$ & 67.87 & 68.09 & 67.48 & 66.49 & \textbf{\color{red}{69.82}} & \\
        \bottomrule
        \end{tabular}}
    \end{sc}
    \end{small}
    \end{center}
    \caption{Ablation test of GWD-based regression loss form and hyperparameter on DOTA. The based detector is RetinaNet.}
    \label{tab:func_tau_}
  %  \vskip -0.2in
\end{table}

\subsection{Improved GWD-based Regression Loss}
In Tab. \ref{tab:func_tau_}, we compare three different forms of GWD-based regression loss, including $d^2$, $1-\frac{1}{\left(\tau+f(d^2)\right)}$ and $f(d^2)$. The performance of directly using GWD ($d^2$) as the regression loss is extremely poor, only 49.11\%, due to its rapid growth trend (as shown on the left of Fig. \ref{fig:loss_form}). In other words, the regression loss $d^2$ is too sensitive to large errors. In contrast, $1-\frac{1}{\left(\tau+f(d^2)\right)}$ achieves a significant improvement by fitting IoU loss. This loss form introduces two new hyperparameters, the non-linear function $f(\cdot)$ to transform the Wasserstein distance, and the constant $\tau$ to modulate the entire loss. From Tab. \ref{tab:func_tau_}, the overall performance of using $sqrt$ outperforms that using $\log$, about 0.98$\pm$0.3\% higher. For $f(\cdot)=sqrt$ with $\tau=2$, the model achieves the best performance, about 68.93\%. In order to further reduce the number of hyperparameters of the loss function, we directly use the GWD after nonlinear transformation ($f(d^2)$) as the regression loss. As shown in the red box in Fig. \ref{fig:loss_form}, $f(d^2)$ still has a nearly linear trend after transformation using the nonlinear function $sqrt$ and only achieves 54.27\%. In comparison, the $\log$ function can better make the $f(d^2)$ change value close to IoU loss (see green box in Fig. \ref{fig:loss_form}) and achieve the highest performance, about 69.82\%. In general, we do not need to strictly fit the IoU loss, and the regression loss should not be sensitive to large errors.

%-------------------------------------------------------------------------

\section*{Acknowledgment}
The author Xue Yang is supported by Wu Wen Jun Honorary Doctoral Scholarship, AI Institute, Shanghai Jiao Tong University. The authors would like to thank Gefan Zhang and Minghuan Liu for their helpful discussion.

{\small
\bibliographystyle{ieee_fullname}
\bibliography{egbib}
}

\end{document}